\documentclass[10pt]{article}

\usepackage{scai}

\usepackage[authoryear, sort&compress, round]{natbib}
\usepackage{xspace}
\usepackage{adjustbox}
\usepackage{array}
\usepackage{float}
\usepackage{dblfloatfix}
\usepackage{setspace}
\usepackage{footmisc}
\usepackage{booktabs}
\usepackage{multirow}
\usepackage{tabularx}
\usepackage{threeparttable}
\usepackage{siunitx}
\usepackage{mathtools}
\usepackage{bm}
\usepackage{dsfont}
\usepackage{subcaption}
\usepackage{tikz}
\usepackage{listings}
\usepackage{cleveref}
\usepackage{fontawesome5}
\usepackage{CJKutf8}
\usepackage{arydshln}
\usepackage{multicol}
\usepackage{mdframed}
\usepackage{wrapfig}
\usepackage[ruled,vlined]{algorithm2e}
\usepackage{nicefrac}
\usepackage{fancyvrb}
\usepackage{fvextra}
\usepackage{sourcecodepro}

\definecolor{nipsbg}{gray}{0.98}
\definecolor{nipsframe}{gray}{0.2}

\newtcolorbox{nipsbox}[1]{
    enhanced,
    breakable,
    sharp corners,
    boxrule=0.5pt,
    colframe=nipsframe,
    colback=nipsbg,
    fonttitle=\bfseries\sffamily\small,
    coltitle=white,
    colbacktitle=nipsframe,
    attach boxed title to top left={yshift=-2mm, xshift=3mm},
    boxed title style={sharp corners, boxrule=0pt, top=0.5mm, bottom=0.5mm, left=2mm, right=2mm},
    title={\texttt{#1}},
    top=4mm, bottom=2mm, left=3mm, right=3mm
}

\title{Unify-Agent: A Unified Multimodal Agent for World-Grounded Image Synthesis}

\contribnote{lead}{These authors share first authorship.}
\contribnote{contrib}{These authors share second authorship.}

\scaiauthor{1,2,lead}{Shawn Chen}
\scaiauthor{4,lead}{Quanxin Shou}
\scaiauthor{2,contrib}{Hangting Chen}
\scaiauthor{2,contrib}{Yucheng Zhou}
\scaiauthor{3,contrib}{Kaituo Feng}
\scaiauthor{1,contrib}{Wenbo Hu}
\scaiauthor{}{Yi-Fan Zhang}
\scaiauthor{3}{Yunlong Lin} 
\scaiauthor{3}{Wenxuan Huang}
\scaiauthor{3}{Mingyang Song}
\scaiauthor{3}{Dasen Dai}
\scaiauthor{4}{Bolin Jiang}
\scaiauthor{3}{Manyuan Zhang}
\scaiauthor{2}{Shi-Xue Zhang}
\scaiauthor{2}{Zhengkai Jiang}
\scaiauthor{2}{Lucas Wang}
\scaiauthor{2}{Zhao Zhong}
\scaiauthor{3}{Yu Cheng}
\scaiauthor{1}{Nanyun Peng}

\affiliation{1}{University of California, Los Angeles}
\affiliation{2}{Tencent Hunyuan} 
\affiliation{3}{The Chinese University of Hong Kong}
\affiliation{4}{The Hong Kong University of Science and Technology}

\metadata[GitHub]{\url{https://github.com/shawn0728/Unify-Agent}}

\begin{abstract}
Unified multimodal models provide a natural and promising architecture for understanding diverse and complex real-world knowledge while generating high-quality images.
However, they still rely primarily on frozen parametric knowledge, which makes them struggle with real-world image generation involving long-tail and knowledge-intensive concepts.
Inspired by the broad success of agents on real-world tasks, we explore agentic modeling to address this limitation. 
Specifically, we present \textbf{Unify-Agent}, a unified multimodal agent for \emph{world-grounded image synthesis}, which reframes image generation as an agentic pipeline consisting of prompt understanding, multimodal evidence searching, grounded recaptioning, and final synthesis.
To train our model, we construct a tailored multimodal data pipeline and curate 143K high-quality agent trajectories for world-grounded image synthesis, enabling effective supervision over the full agentic generation process. 
We further introduce FactIP, a benchmark covering 12 categories of culturally significant and long-tail factual concepts that explicitly requires external knowledge grounding. 
Extensive experiments show that our proposed Unify-Agent substantially improves over its base unified model across diverse benchmarks and real world generation tasks, while approaching the world knowledge capabilities of the strongest closed-source models.
As an early exploration of agent-based modeling for world-grounded image synthesis, our work highlights the value of tightly coupling reasoning, searching, and generation for reliable open-world agentic image synthesis.
    
\end{abstract}

\begin{document}
\maketitle

\section{Introduction}
\label{sec:intro}

\begin{figure}[htbp]
    \centering
    \includegraphics[width=1.00\textwidth,height=0.95\textheight,keepaspectratio]{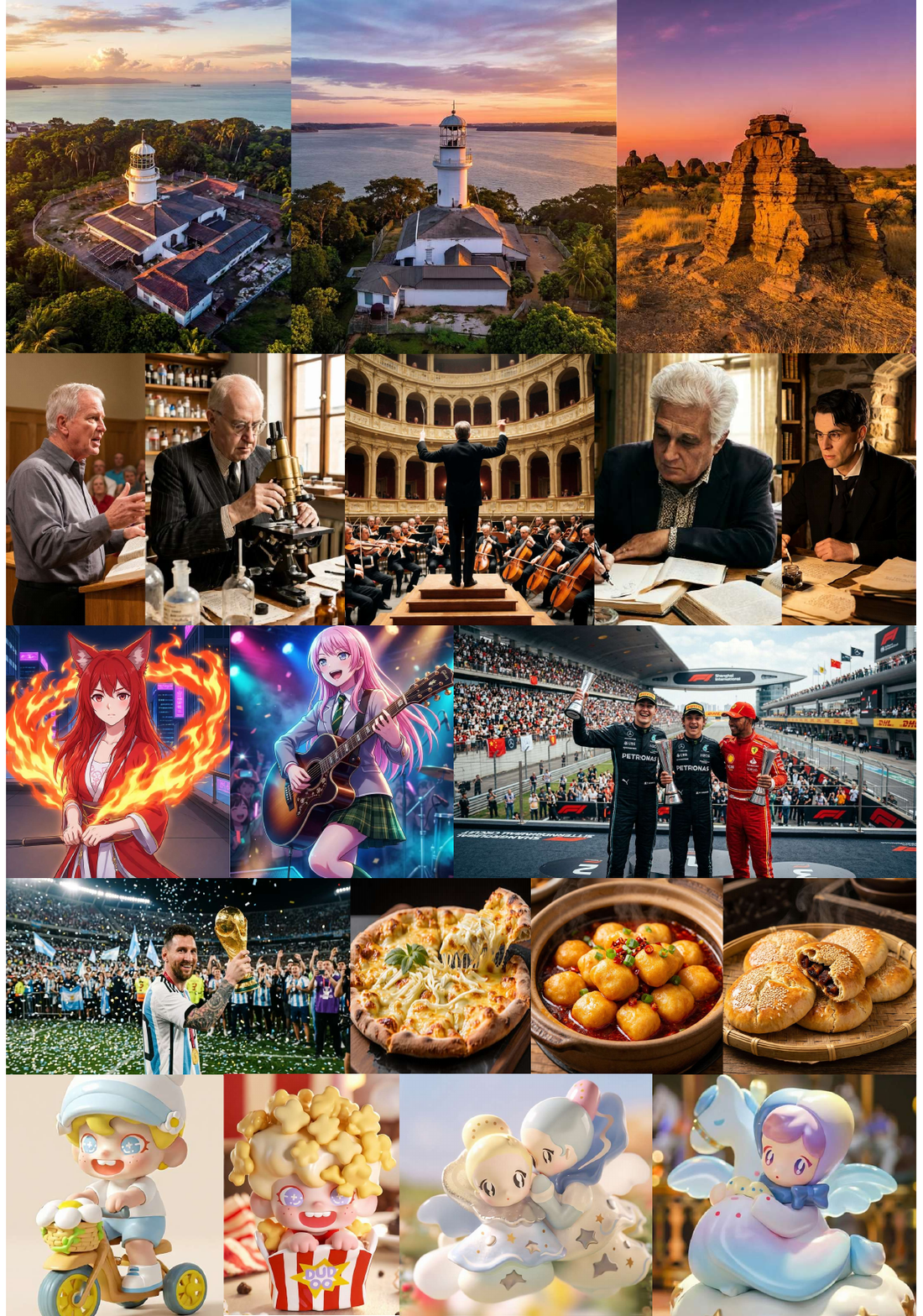}
    \caption{High-quality samples from our \textbf{Unify-Agent}, highlighting its excellence in unified multi-image generation and agentic search enhanced world knowledge integration. It delivers strong cross-image consistency, broad stylistic versatility, and more faithful, knowledge-grounded visual generation across diverse concepts and scenarios.}
    \label{fig:showcase}
\end{figure}

Recent advances in Text-to-Image (T2I) generation has significantly improved visual realism, controllability, and stylistic diversity~\citep{rombach2022high, esser2024scaling, sun2024autoregressivemodelbeatsdiffusion, flux1dev2025}. Despite these developments, visual quality alone remains insufficient for many practical applications. A growing challenge is whether such models can faithfully depict entities and concepts grounded in the real world, including real people, cultural symbols, rare intellectual properties (IPs), historical scenes, scientific phenomena, and other long-tail concepts. In such cases, success requires more than visually plausible outputs, it also requires factual and visual fidelity to the intended entity. Unified multimodal models (UMMs) offer a promising approach to this challenge by unifying visual understanding and image generation within a shared architecture, allowing world knowledge and multimodal reasoning to inform image synthesis~\citep{team2024chameleon, xie2024showo, wu2024janus, deng2025bagel, zhou2024transfusion}. Existing models are limited to parametric knowledge, which is often insufficient to recover the correct appearance and identity-defining visual cues of the target. This limitation reflects a fundamental constraint: the world knowledge is embedded in fixed model parameters and cannot be updated at inference time. Therefore, factual image synthesis may require external world knowledge at inference time rather than sole reliance on parametric memory, as the key challenge is not rendering quality but knowing the target’s correct appearance and identity-defining visual cues.

We argue that addressing this bottleneck requires moving beyond \emph{closed-book generation} toward \emph{open-book, agentic generation}, where models can access external world knowledge at inference time instead of relying solely on parametric memory. Yet existing agentic text-to-image systems fall short of this vision. Most are brittle, multi-stage pipelines that loosely connect an LLM planner, retrieval tools, and a standalone image generator~\citep{wang2024genartist, chen2025t2icopilot, son2025worldtoimage, he2026mindbrush}. Consequently, evidence acquisition, multimodal reasoning, and visual synthesis remain separate stages rather than parts of a unified generative process, making effective evidence integration difficult. Retrieved text can enrich semantic context, but it rarely specifies the fine-grained visual attributes needed for faithful generation. Reference images, although visually informative, often include irrelevant background elements and may conflict with the user-specified composition. Therefore, the key challenge is not simply retrieving external world knowledge. It is turning that knowledge into visual guidance that can both preserve fidelity to the source knowledge and remain aligned with user intent. This capability remains missing in current agentic image generation systems.

To this end, we present Unify-Agent, the first end-to-end unified multimodal agent for world-grounded image synthesis, to the best of our knowledge. Unify-Agent integrates four tightly coupled capabilities within a unified multimodal model: (1) \emph{THINK}, for prompt understanding and information-gap detection; (2) \emph{RESEARCH}, for actively acquiring relevant textual and visual knowledge from external sources; (3) \emph{RECAPTION}, for converting retrieved knowledge into structured textual guidance; and (4) \emph{GENERATE}, for final image synthesis. Rather than passing raw retrieved content to the generator, Unify-Agent uses recaptioning as an intelligent filtering mechanism to transform external knowledge into structured textual constraints, explicitly disentangling identity-preserving cues from scene-compositional requirements.

To realize this vision in practice, we build the full agentic workflow on a unified model instead of a decoupled pipeline. This architectural choice is motivated by a key insight: \textbf{generative priors may offer a promising way to enhance multimodal understanding.} In decoupled systems, language-based reasoning often overlooks fine-grained visual attributes, limiting the interpretation of retrieved visual references. In contrast, Unify-Agent leverages recaptioning as a natural bridge between understanding and generation. The Vision Transformer (ViT) captures high-level semantics, including global context and entity identity, while the Variational Autoencoder (VAE), although originally designed for image synthesis, provides low-level perceptual latents that retain details such as texture, material, and structural geometry. Together, these components enable Unify-Agent to interpret retrieved visual references with higher fidelity and convert them into precise textual specifications for generation. To evaluate this capability, we introduce FactIP, a comprehensive benchmark of 2,462 curated prompts centered on rare identities and culturally significant long-tail concepts. Experimental results show that Unify-Agent significantly improves factual visual synthesis. On FactIP, it attains an overall score of 73.2, surpassing its base model (Bagel) by more than 22 points and outperforming strong generation-only baselines such as FLUX.1-dev and SD-3.5-large. Moreover, Unify-Agent achieves superior performance among open-source unified models on several widely used world-knowledge benchmarks, including WiSE, KiTTEN, and T2I-FactualBench, while also providing a promising direction for enhancing world knowledge in unified models.

In summary, our main contributions are as follows:
\begin{itemize}[leftmargin=*, noitemsep]
  \renewcommand\labelitemi{$\diamond$}
    \item \textbf{A Novel Agentic Paradigm.} We introduce Unify-Agent, the first end-to-end unified multimodal agent that reformulates text-to-image generation from a passive mapping into an active, inference-time sequential decision process (Think, Research, Recaption, Generate).
    \item \textbf{Architectural Insights on UMMs.} We reveal that unifying understanding and generation creates a mutually reinforcing synergy. Specifically, we demonstrate that the joint availability of low-level generative latents (VAE) and high-level semantic tokens (ViT) enables vastly superior multimodal reasoning during evidence recaptioning.
    \item \textbf{A Comprehensive Benchmark.} We present FactIP, a highly curated benchmark designed specifically to evaluate the identity consistency and factual faithfulness of world-grounded generation for rare and long-tail concepts.
    \item \textbf{Superior Performance.} Unify-Agent sets superior records among open-source unified models across several established factual benchmarks (FactIP, WiSE, KiTTEN, and T2I-FactualBench), seamlessly blending real-world knowledge with creative visual synthesis, and demonstrates world knowledge capabilities approaching those of leading commercial models.
\end{itemize}

\section{Related Works}

\subsection{Unified multimodal models} 

Recent unified multimodal models (UMMs)~\citep{team2024chameleon, zhou2024transfusion, deng2025bagel, guo2025seed15vltechnicalreport, huang2025interleavingreasoningbettertexttoimage} aim to jointly support visual understanding and image generation within a shared backbone. Architectures like the Janus family~\citep{wu2024janus, chen2025januspro} and Show-o~\citep{xie2024showo} have demonstrated that semantic comprehension and continuous visual synthesis can be effectively harmonized. Despite this, current UMMs remain strictly \emph{closed-book} systems. Because they rely exclusively on static parametric memory internalized during pre-training, they frequently hallucinate or suffer from identity drift when prompted with rare, long-tail, or world-dependent entities~\citep{huang2025t2ifactualbench, chen2024mllm, chen2026surveymultimodalhallucinationevaluation}. This bottleneck highlights the need to shift from purely parametric generation toward an \emph{open-book} approach, where the model actively acquires external evidence before synthesis.

\subsection{Agentic workflows} 

As reasoning in large language models~\citep{Guo_2025, chen2025advancing, chen2025ares, feng2025onethinkerallinonereasoningmodel} is increasingly evolving into agentic reasoning~\citep{yao2023react,zeng2026vision,huang2026vision, shou2026halo, fan2026exploringreasoningrewardmodel}, the focus is shifting from isolated deliberation to action-oriented planning with tools and environmental interaction. In line with this transition, recent efforts in text-to-image generation~\citep{wang2024genartist, chen2025t2icopilot, son2025worldtoimage} have attempted to incorporate similar agentic pipelines. However, these approaches predominantly adopt a fragile \emph{multi-API stitching} paradigm~\citep{shen2024hugginggpt, wu2023visualchatgpt, yang2023autogpt, yang2023mmreactpromptingchatgptmultimodal, he2026mindbrush}. They loosely chain together frozen text-only planners, external search tools, and standalone image generators~\citep{shen2024hugginggpt, wu2023visualchatgpt, feng2025gensearcher}, failing to realize a true end-to-end UMM agentic framework. This shallow integration inevitably introduces cascading errors and rigidly separates multimodal reasoning from visual synthesis. In contrast, our Unify-Agent pioneers an \emph{end-to-end unified agentic framework}. By natively consolidating cognitive gap detection, cross-modal evidence grounding, and generation planning within a single cohesive architecture, we eliminate the reliance on brittle API pipelines and achieve genuine reasoning-driven generation.

\subsection{World knowledge and factual evaluation.} 
Conventional text-to-image benchmarks predominantly emphasize aesthetic quality and generic prompt alignment~\citep{huang2023t2icompbench, lee2024heim}, largely overlooking the factual correctness of generated content. To address this, recent efforts have pivoted towards knowledge-oriented evaluation. Notably, T2I-FactualBench~\citep{huang2025t2ifactualbench} provides a comprehensive diagnosis of generative models, systematically exposing their struggles with knowledge-intensive concepts across multiple domains. This severe deficiency in explicit world knowledge is further echoed by related benchmarks such as KITTEN~\citep{huang2024kitten} and WISE~\citep{niu2025wise}. Nevertheless, existing evaluations remain fundamentally \emph{diagnostic}---they quantify the knowledge gap but offer no structural remedy. Motivated by these insights, we not only construct a targeted evaluation for rare identities and long-tail concepts, but also propose an evidence-grounded agentic framework that actively bridges this gap during synthesis.

\section{Preliminary}

\subsection{Unified Multimodal Model: Bagel}
\label{pril:bagel}

We build our Unify-Agent upon Bagel~\citep{deng2025emergingbagel}, leveraging its native capabilities to develop a unified and interleaved reasoning framework for world-grounded image synthesis. At its core, Bagel employs a Mixture-of-Transformers (MoT) architecture, integrating a ViT encoder~\citep{tschannen2025siglip} to process multimodal inputs. This architectural design gracefully unifies visual understanding and continuous image generation within a single foundation model. Specifically, the model disentangles these two core capabilities through dedicated experts:

\textbf{Multimodal Understanding.} The understanding pathway is formulated as an autoregressive next-token prediction task. Handled by a dedicated understanding expert, the model generates context-aware textual responses via a language modeling head. Conditioned on the multimodal input context $C$, the training objective minimizes the negative log-likelihood:
\begin{equation}
\mathcal{L}_{\text{text}} = - \sum_{t=1}^{T} \log p_{\theta}\!\left(x_t \mid x_{<t}, C\right),
\end{equation}
where $x_t$ is the target text token, $x_{<t}$ denotes the preceding token sequence, and $C$ encapsulates the provided multimodal context.

\textbf{Multimodal Generation.} Conversely, the generation pathway is designed to synthesize high-fidelity, semantically aligned images. Handled by a generation expert, this process is formulated as a rectified flow~\citep{liu2022flowstraightfastlearning} operating within the latent space of a continuous VAE~\citep{labs2025flux1kontextflowmatching}. Conditioned on the same multimodal context $C$, the model learns a time-conditioned velocity field $u_{\theta}$ by minimizing the latent flow-matching objective:
\begin{equation}
\mathcal{L}_{\text{image}} = \mathbb{E}_{t \sim \mathcal{U}(0,1), z_t} \left\| u_{\theta}\!\left(z_t, t \,;\,C\right) - u^{\star}\!\left(z_t, t\right) \right\|_2^2,
\end{equation}
where $t \sim \mathcal{U}(0,1)$ is the continuous timestep, $z_t$ represents the latent state at time $t$, $u^{\star}(z_t, t)$ is the target vector field, and $u_{\theta}(z_t, t; C)$ is the velocity field predicted by the generation expert.

\subsection{Motivating Evidence for World-Grounded Synthesis}
\label{sec:motivation}

A central challenge in world-grounded image synthesis is that failures on out-of-distribution concepts, especially rare intellectual properties (IPs), do not primarily stem from insufficient visual fidelity, but from missing world knowledge. When prompted with a long-tail character, scene, or object, a base T2I model often does not know what the concept should look like, which attributes are identity-defining, or how those attributes should be instantiated under compositional instructions. This observation suggests that improving factual generation for rare IPs requires augmenting the model with external knowledge at inference time. Intuitively, textual knowledge injection can supplement semantic definitions, background context, and key attributes, while visual knowledge injection can provide direct appearance anchors that stabilize identity and concept realization.

\begin{wrapfigure}{r}{0.5\textwidth}
    \vspace{-10pt} 
    \centering
    \includegraphics[width=0.5\textwidth]{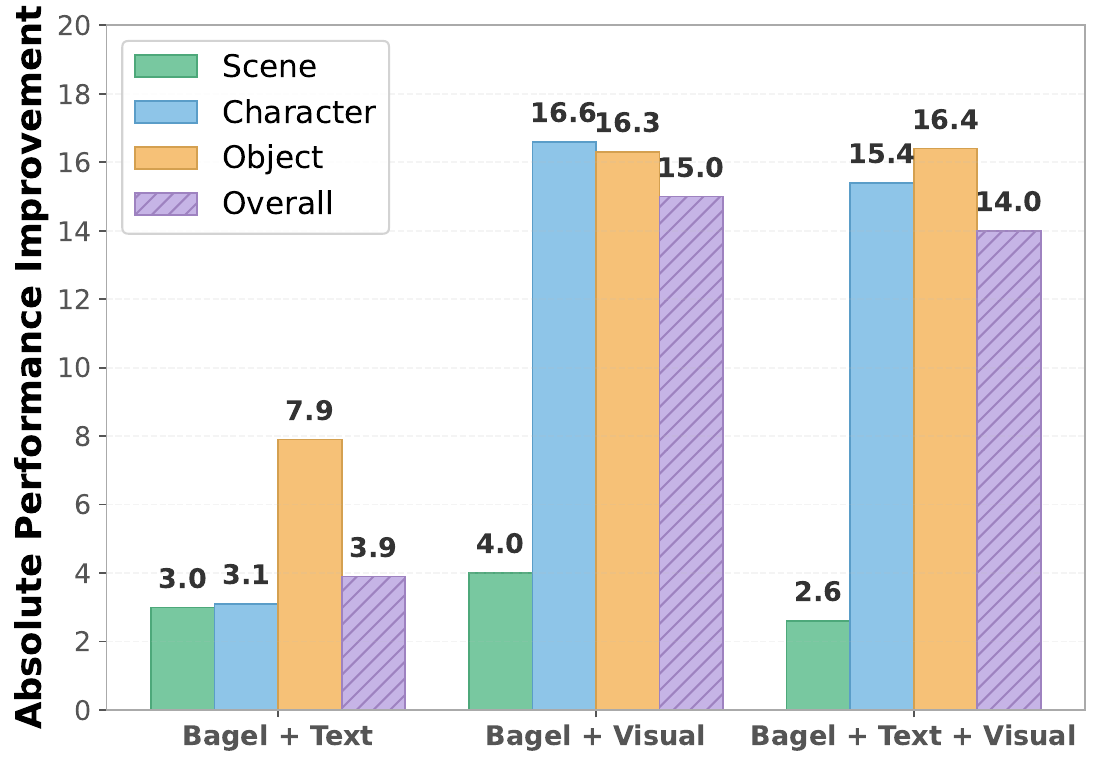}
    \caption{Absolute performance improvement over prompt-only \textbf{Bagel} on rare-IP factual generation under different knowledge injection strategies. We compare text injection, visual injection, and joint text+visual injection across three FactIP categories: Scene, Character, and Object. The purple hatched bars denote the overall average gain across all evaluated samples.
          }
    \label{fig:prili_bar}
    \vspace{-10pt}  
\end{wrapfigure}

To validate this intuition, we conduct a preliminary training-free study on 200 examples sampled from three major categories in FactIP Benchmark: \textit{scene}, \textit{character}, and \textit{object}. Using Bagel as the base model, we compare four inference settings: (1) direct prompting with Bagel only, (2) Bagel with \textbf{text injection}, where IP-related background information is appended to the original prompt, (3) Bagel with \textbf{visual injection}, where two ground-truth reference images of the target IP are provided, and (4) Bagel with \textbf{text + visual injection}, where both sources of external knowledge are injected simultaneously. As shown in Figure~\ref{fig:prili_bar}, both textual and visual knowledge consistently improve over the prompt-only baseline, confirming that external knowledge is beneficial for factual synthesis. More importantly, visual injection yields substantially larger gains. These results further suggest that enabling a model to proactively search for external knowledge during inference can directly improve its reasoning and generation for rare IP concepts.

At the same time, Figure~\ref{fig:prili_bar} also reveals the limitation of naive knowledge injection. Although combining text and visual inputs remains beneficial overall, its improvement is slightly weaker than visual injection alone, suggesting that simply concatenating retrieved text with prompt instructions does not optimally exploit multimodal evidence. Raw text injection may introduce redundant or weakly visual information, which can interfere with instruction following under complex prompts; conversely, raw visual injection, while effective at anchoring identity, may over-constrain the generation process and reduce flexibility in attribute manipulation or compositional reasoning. These findings motivate our next step: rather than directly injecting external knowledge in its raw textual or visual form, we seek a better interface that can transform multimodal evidence into a representation more suitable for generative modeling. This observation directly paves the way for our recaption paradigm, which distills retrieved textual and visual knowledge into a unified, structured, and generation-oriented description for world-grounded synthesis.

\subsection{Problem Formulation}
\label{sec:problem_formulation}

Standard T2I generation models the conditional distribution $p_{\theta}(y \mid x)$, where an image $y \in \mathcal{Y}$ is synthesized from a user prompt $x \in \mathcal{X}$ relying strictly on the parametric memory $\theta$. However, for world-grounded synthesis involving rare entities, estimating $p_{\theta}(y \mid x)$ directly is intractable due to knowledge deficits in $\theta$. Motivated by the empirical findings in Section~\ref{sec:motivation}, we recognize that naively expanding the condition to $p_{\theta}(y \mid x, \mathcal{K}_{\text{text}}, \mathcal{K}_{\text{vis}})$, where $\mathcal{K}$ represents raw retrieved external knowledge, leads to suboptimal alignment. Raw multimodal evidence introduces formatting noise and conflicting constraints that disrupt the continuous generative process.

Consequently, we formulate world-grounded image synthesis not as a single-step mapping, but as an \textbf{interleaved generative trajectory} over an augmented state space. We introduce four intermediate variables to bridge the gap between prompt understanding and visual synthesis: the cognitive gap assessment $g$, the textual evidence trace $\tau_t$, the visual evidence trace $\tau_v$, and the evidence-grounded recaption $c$. The joint distribution thus defines the holistic generative process:
\begin{align}
\label{eq:joint_prob}
p_{\theta}(y, c, \tau_t, \tau_v, g \mid x) = \underbrace{p_{\theta}(g \mid x)}_{\text{Gap Detection}} \cdot \underbrace{p_{\theta}(\tau_t, \tau_v \mid x, g)}_{\text{Evidence Acquisition}} \cdot \underbrace{p_{\theta}(c \mid x, g, \tau_t, \tau_v)}_{\text{Evidence-Grounded Recaptioning}} \cdot \underbrace{p_{\theta}(y \mid c, \tau_v)}_{\text{Visual Synthesis}}
\end{align}
Optimizing Equation~\ref{eq:joint_prob} requires unifying discrete sequence modeling (for $g, \tau_t, \tau_v,$ and $c$) and continuous latent modeling (for $y$) within a shared parameter space $\theta$. Under this unified paradigm, the probabilistic decomposition rigorously defines the four distinct cognitive phases of \textbf{Unify-Agent}:
\begin{enumerate}[leftmargin=*]
    \item \textbf{Cognitive Gap Detection} ($p_{\theta}(g \mid x)$): The model first evaluates the prompt $x$ against its internal parametric memory to identify missing visually-critical attributes, formulating an internal reasoning trace $g$ to decide whether external world knowledge is required.
    \item \textbf{Evidence Acquisition} ($p_{\theta}(\tau_t, \tau_v \mid x, g)$): Conditioned on the identified knowledge gap $g$, the agent proactively interacts with external environments to search and reason over textual evidence ($\tau_t$), and subsequently acquires and reasons over visual evidence ($\tau_v$).
    \item \textbf{Evidence-Grounded Recaptioning} ($p_{\theta}(c \mid x, g, \tau_t, \tau_v)$): To resolve the modality conflict of raw injection, the model consolidates gathered evidence and the original instruction into a highly structured, generation-oriented recaption $c$.
    \item \textbf{Visual Synthesis} ($p_{\theta}(y \mid c, \tau_v)$): The final image generation is conditioned exclusively on the recaption $c$ and the visual identity anchors $\tau_v$. By formally enforcing conditional independence from the noisy reasoning history $x, g,$ and $\tau_t$, this step ensures that the generative prior is strictly guided by refined constraints.
\end{enumerate}

\section{Data Pipeline}
\label{sec:data}

\begin{figure}[t]
    \centering
    \includegraphics[width=1.00\textwidth,keepaspectratio]{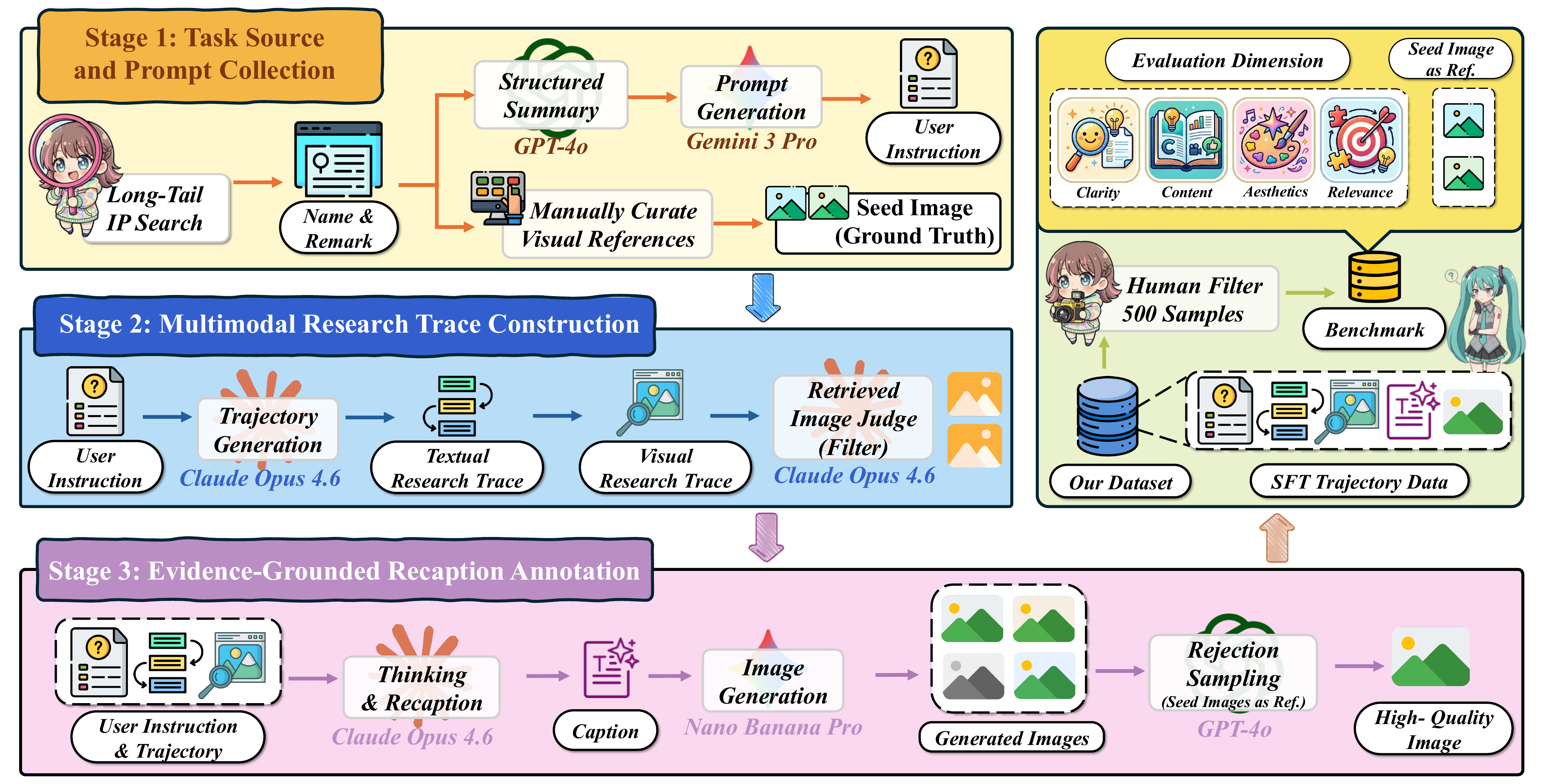}
    \caption{\textbf{Overview of our data pipeline.}
Starting from long-tail IP collection, we construct user instructions and Ground Truth images, build multimodal research trajectories with textual and visual evidence, and finally perform evidence-grounded recaption annotation to obtain high-quality training samples. The resulting dataset supports both SFT trajectory learning and the FactIP benchmark, which evaluates generation quality in terms of clarity, content, aesthetics, and relevance.}
    \label{fig:data_pipeline}
\end{figure}

To train a unified multimodal agent for world-grounded image synthesis, we require data that captures not only the final user instruction and output, but also the intermediate process of multimodal reasoning, evidence searching, and evidence aggregation. Meanwhile, to evaluate whether the model can generate images grounded in factual and long-tail world knowledge, we construct a dedicated benchmark. As illustrated in Figure~\ref{fig:data_pipeline}, our overall data pipeline therefore comprises two complementary components: (1) training data construction for agent supervised fine-tuning, and (2) a curated FactIP evaluation benchmark. In the following sections, we provide a detailed illustration of the data construction pipeline and strategies.

\subsection{Training Data Construction}
\label{sec:traing_data_curation}

To train Unify-Agent to execute the full \textsc{Think}--\textsc{Research}--\textsc{Generate} pipeline, we construct supervision that covers not only the final generation-oriented recaption, but also the intermediate evidence-acquisition process. 
Accordingly, each training sample is represented as
\begin{equation}
\mathcal{D}_{\text{SFT}} = \{(x, \tau_t, \tau_v, c)\},
\end{equation}
where $x$ denotes the original user prompt, $\tau_t$ denotes the textual research trace, $\tau_v$ denotes the visual research trace, and $c$ denotes the final evidence-grounded recaption. 
Under this formulation, $\tau_t$ and $\tau_v$ supervise how the agent performs external evidence acquisition, while $c$ supervises how the acquired evidence is transformed into a synthesis-ready specification.

Compared with standard instruction tuning for multimodal models, our data differs in two important respects. 
First, the target output is not a generic assistant response or image caption, but a structured recaption designed to preserve identity-critical visual attributes while remaining faithful to user-specified scene and style constraints. 
Second, the supervision is process-aware: instead of learning only the final output, the model is trained on explicit multimodal research traces, enabling it to internalize not only \emph{what} evidence is useful, but also \emph{when} and \emph{how} such evidence should be acquired before generation.

We construct the training corpus in three stages: (1) task source curation and prompt collection, (2) multimodal research trace construction, and (3) evidence-grounded recaption annotation with generation-based verification.

\subsubsection{Task Source and Prompt Collection}
\label{sec4.1.1 task_source}

\paragraph{Task Source.}
We begin by constructing a large-scale pool of knowledge-intensive IPs that are likely to expose world-knowledge deficiency in image generation models. To this end, we organize a large group of annotators to broadly collect relatively long-tail and less frequently represented IPs from the web, rather than relying on only popular or highly recurrent entities. The collected IPs are categorized into 12 domains, including \textit{Celebrity}, \textit{Animation}, \textit{Game}, \textit{Comic}, \textit{Mythology}, \textit{Mascot}, \textit{Animal}, \textit{Food}, \textit{Art}, \textit{Toy}, \textit{Landmark}, and \textit{Festival}. For each concept, we query BabelNet \citep{navigli-ponzetto-2010-babelnet} and Wikipedia to retrieve relevant webpage information, and annotators further search for two representative seed images as the ground-truth visual references. We then use \textit{GPT-4o} \citep{openai2024gpt4ocard} to summarize each IP into structured metadata, including its identity-defining attributes and relevant factual descriptions. To ensure data reliability, we perform an additional round of manual verification to remove samples with incorrect metadata, low-quality seed images (e.g., severe watermarks or blurry content), or concepts that are overly common and do not sufficiently challenge factual grounding. After this filtering process, we obtain a final dataset containing \textbf{456K} examples.

\paragraph{Prompt Collection.}
Based on the curated IP pool, we construct user instructions designed to evaluate whether a model can faithfully preserve grounded identity while performing creative image generation. Rather than using arbitrary prompts, we generate prompts that retain the original generation intent, such as scene composition, clothing or object specification, atmosphere, and photographic or cinematic style, so that the task requires both factual consistency and compositional flexibility. In particular, for different categories, we use \textit{Gemini 3 Pro}~\citep{geminiteam2025geminifamilyhighlycapable} to create diverse prompts tailored to their domain characteristics. These prompts cover a spectrum of difficulty, ranging from direct rendering of a target identity or concept to more challenging cases where the model must preserve defining attributes while adapting pose, costume, scene, lighting, or aesthetic style. Such a construction improves diversity across tasks and helps avoid overfitting to simple retrieval-like behavior, encouraging the model to handle a broader range of grounded generation scenarios.

\subsubsection{Multimodal Research Trace Construction}
\label{sec:data_mm_search}

To instantiate the evidence acquisition process in Eq.~\ref{eq:joint_prob}, we construct supervised multimodal research traces that approximate the desired inference-time behavior of the agent. 
For each prompt $x$ collected from Section~\ref{sec4.1.1 task_source}, we use a strong teacher agent to externalize the intermediate evidence-seeking process and serialize it into a structured trajectory. 
Concretely, we employ \textit{Claude Opus 4.6} as the trajectory-construction model to perform multi-step research over external tools and to produce high-quality agent traces. 
This choice is motivated by its strong capability on long-horizon agentic tasks, tool use, and complex research-style workflows, making it well-suited for synthesizing supervision signals that reflect realistic evidence acquisition behavior.

Given a prompt $x$, trajectory construction is formulated as a \emph{sequential evidence acquisition process} with two stages: \emph{textual research} followed by \emph{visual research}. 
This ordering is deliberate. 
Textual evidence first resolves semantic ambiguity and establishes high-level grounding, which in turn enables more precise and context-aware visual search.

\paragraph{Textual research trace.}
The textual research trace $\tau_t$ captures how the agent acquires semantic grounding from external knowledge sources. 
Conditioned on the prompt $x$ and the inferred cognitive gap $g$, the teacher agent first formulates a textual query $q_t$:
\begin{equation}
q_t \sim p_{\theta}(q_t \mid x, g),
\end{equation}
which is issued to an external text searching system to obtain textual evidence
\begin{equation}
E_t = \mathrm{Retrieve}_{\mathrm{text}}(q_t).
\end{equation}

The role of $E_t$ is not to directly replace the user prompt, but to provide a compact semantic scaffold for downstream reasoning. 
In practice, $E_t$ typically contains information such as identity disambiguation, categorical background, role-specific context, and other descriptors that are useful for constraining subsequent visual search. 
We record both the issued query $q_t$ and the resulting evidence summary $E_t$ as part of the supervised trace, thereby teaching the model not only \emph{what} knowledge to use, but also \emph{how} to acquire it.

\paragraph{Visual research trace.}
After establishing textual grounding, the agent proceeds to visual evidence acquisition. 
Conditioned on $(x, g, \tau_t)$, the teacher agent generates a visual query $q_v$:
\begin{equation}
q_v \sim p_{\theta}(q_v \mid x, g, \tau_t),
\end{equation}
where $q_v$ is designed to retrieve images that are both identity-preserving and context-compatible. 
Compared with naive retrieval based solely on entity names, this query incorporates both prompt-specific constraints and the semantic grounding provided by $\tau_t$, thereby enabling more precise and scenario-aware visual search.

The initial candidate set is retrieved as
\begin{equation}
\widetilde{E}_v = \mathrm{Retrieve}_{\mathrm{image}}(q_v)
= \{v_1, \dots, v_n\}.
\end{equation}
To identify high-quality visual references, we further employ \textit{Gemini 3 Flash} \citep{geminiteam2025geminifamilyhighlycapable} as a lightweight visual evaluator to score each candidate along four dimensions: \textbf{identity consistency}, \textbf{subject salience}, \textbf{image clarity}, and \textbf{watermark cleanliness}. 
Concretely, a preferred image should faithfully match the target identity or concept, present the target as the dominant subject, maintain sufficient visual quality, and avoid heavy watermarks or text-dominant compositions.
The overall score for each candidate is computed as
\begin{equation}
s(v_i) = \sum_{k=1}^{4} \lambda_k \, s_k(v_i \mid x, E_t).
\end{equation}

Based on these scores, the candidate pool is ranked and the top two images are selected as the final visual evidence:
\begin{equation}
E_v = \operatorname*{arg\,top2}_{v_i \in \widetilde{E}_v} \; s(v_i).
\end{equation}
The resulting images serve as the primary visual anchors for subsequent recaptioning and synthesis. 
By supervising both the query formulation process and the final selected references, the model learns visual grounding as a structured evidence acquisition procedure, rather than simply consuming pre-selected images.

\paragraph{Trace representation and supervision.}
The final multimodal research trace is represented as
\begin{equation}
(\tau_t, \tau_v) = \big(q_t, E_t, q_v, E_v\big),
\end{equation}
where $q_t$ and $q_v$ denote the textual and visual queries, $E_t$ denotes the retrieved textual evidence, and $E_v$ denotes the selected visual evidence. 
This representation makes the evidence acquisition process explicit and directly aligns with the factorized term $p_{\theta}(\tau_t, \tau_v \mid x, g)$ in Eq.~\ref{eq:joint_prob}.

By supervising these intermediate trajectories, we train the model to treat world grounding as a structured, multi-step process of external evidence construction. 
Rather than collapsing grounding into a single retrieval-augmented prompt, the model learns to actively formulate queries, gather heterogeneous evidence, and organize that evidence into a form suitable for downstream recaptioning and image synthesis.

\subsubsection{Evidence-Grounded Recaption Annotation}
\label{sec:data_recaption}

As discussed in Section~\ref{sec:motivation}, directly injecting externally retrieved evidence into the original user instruction is not an optimal interface for world-grounded image synthesis. 
Although both textual and visual evidence can improve factual generation, naive multimodal injection remains suboptimal: raw text often introduces redundant or weakly visual information that interferes with instruction following, while raw visual inputs alone may over-constrain the generation process and reduce compositional flexibility. 
This observation suggests that external evidence should not be consumed in its original form, but instead transformed into a representation that is more compatible with downstream generative modeling.

Motivated by recent advances in prompt rewriting and model-facing specification design for image generation, as exemplified by DALL-E 3 \citep{betker2023improving}, PromptEnhancer \citep{promptenhancer}, and HunyuanImage 3.0 \citep{hunyuanimage30}, we adopt an \emph{evidence-grounded recaption} as the core intermediate supervision target. 
This recaption is not a generic image caption and not a simple paraphrase of the prompt. 
Instead, it is a structured textual specification intended to control downstream image synthesis. 
Its role is to consolidate the original instruction with retrieved external evidence into a generation-oriented description that preserves identity-critical attributes while maintaining scene-level and stylistic controllability.

Formally, given the user prompt $x$ and the retrieved multimodal evidence $(E_t, E_v)$, we construct a grounded recaption
\begin{equation}
c = \mathcal{C}(x, E_t, E_v),
\end{equation}
where $\mathcal{C}$ denotes the recaptioning function. 
The resulting recaption integrates three complementary sources of information: (1) prompt-level compositional constraints such as scene, pose, clothing, atmosphere, and rendering style; (2) semantic context derived from textual evidence for identity disambiguation and background consistency; and (3) identity-preserving visual cues grounded in the selected reference images. 
In this way, recaptioning serves as a structured interface that translates heterogeneous external evidence into a unified textual control signal suitable for generation.

To ensure that the annotated recaptions are not only semantically grounded but also operationally useful for image synthesis, we further perform a generation-based validation procedure. 
Specifically, after obtaining the grounded recaption $c$, we feed it together with the two selected reference images $E_v=\{v_1, v_2\}$ into \textit{Nano Banana Pro} \citep{geminiteam2025geminifamilyhighlycapable} to synthesize an image
\begin{equation}
\hat{y} \sim p_{\phi}(y \mid c, E_v),
\end{equation}
where $p_{\phi}$ denotes the image generator used for data construction. 
The generated image $\hat{y}$ is then compared against the ground-truth image of the corresponding IP instance using \textit{GPT-4o} \citep{openai2024gpt4ocard} as a multimodal judge, which evaluates whether the synthesized result is sufficiently faithful to the intended identity and visual concept.

Based on the judge outcome, we adopt a reject-sampling~\citep{liu2024statisticalrejectionsamplingimproves} strategy to improve annotation reliability. 
If the generated image fails the identity-consistency check, we re-run the recaption-to-generation process and repeat the validation step, for up to five trials in total. 
If no satisfactory result is obtained after five attempts, we treat the failure as evidence that the underlying trajectory is unreliable---typically due to erroneous retrieval, weak reference quality, or mismatched visual anchors---and discard the entire trajectory from the training set. 
This process effectively filters out noisy supervision signals that would otherwise teach the model incorrect grounding behavior.

After this recaptioning-and-verification pipeline, we obtain a final set of \textbf{143K high-quality trajectory-image pairs} for subsequent training. 
These samples provide supervision not only over the final synthesis target, but also over the complete evidence-grounded reasoning path that leads to it. 
As a result, the model is trained to view world-grounded generation as a multi-stage process: acquire external evidence, reorganize it into a synthesis-ready recaption, and generate images that remain faithful to both the retrieved world knowledge and the user's compositional intent.

\subsection{FactIP Evaluation Benchmark}
\label{sec: data_factip_benchmark}

\subsubsection{Benchmark Construction}

Our \textbf{FactIP} benchmark is derived from the task pool constructed in Section~\ref{sec4.1.1 task_source}, but is strictly separated from the training data. Specifically, after collecting long-tail IP candidates and organizing them into user-facing generation tasks, we reserve a disjoint subset for evaluation to ensure that benchmark instances are not used during supervised fine-tuning. Each benchmark sample contains a user instruction, two seed images (used as ground truth), and the necessary metadata associated with the target IP.

We design FactIP to evaluate world-grounded image synthesis under challenging yet realistic settings. In particular, benchmark samples are selected to satisfy four principles: (1) \textbf{long-tail and knowledge-intensive}, such that successful generation requires external or internalized world knowledge beyond common popular concepts; (2) \textbf{visually grounded}, such that each target is associated with a clear visual identity that can be assessed against reference images; (3) \textbf{difficult for memorization-only models}, such that strong performance cannot be achieved by relying solely on pretraining memorization or generic text-to-image priors; and (4) \textbf{diverse across categories}, so that the benchmark covers a broad range of entities and scenarios.

Starting from a larger candidate pool, we first obtain 2,500 samples through manual filtering. We then construct a lightweight test split of 500 samples by further sampling with the same category distribution. During this process, we remove samples with severe information noise, unclear instructions, low-quality reference images, or highly ambiguous identities. This filtering procedure ensures that each retained sample exhibits a clear visual identity and a genuine demand for factual grounding, yielding a high-quality benchmark for evaluating world-grounded generation. Additional benchmark construction details are provided in Appendix~\ref{appendix: factip_construction}.

\subsubsection{Evaluation Criterion}

We evaluate generated images on FactIP using a structured multi-dimensional protocol. In particular, each sample is assessed with the user instruction and the Ground Truth (seed) image as reference, which helps determine both \textit{identity consistency} and \textit{factual faithfulness}. Our evaluation considers four dimensions: \textbf{Clarity}, \textbf{Content}, \textbf{Aesthetics}, and \textbf{Relevance}. Among them, \textit{Relevance} is especially important, as it measures whether the generated result preserves the defining identity cues of the target IP while remaining consistent with the instruction and reference image. For brevity, we defer the full evaluation protocol, scoring details, and weighting scheme to Appendix~\ref{appendix: factip_eval}.

\section{Methodology}
\label{sec:method}

\begin{figure}[t]
    \centering
    \includegraphics[width=1.00\textwidth,keepaspectratio]{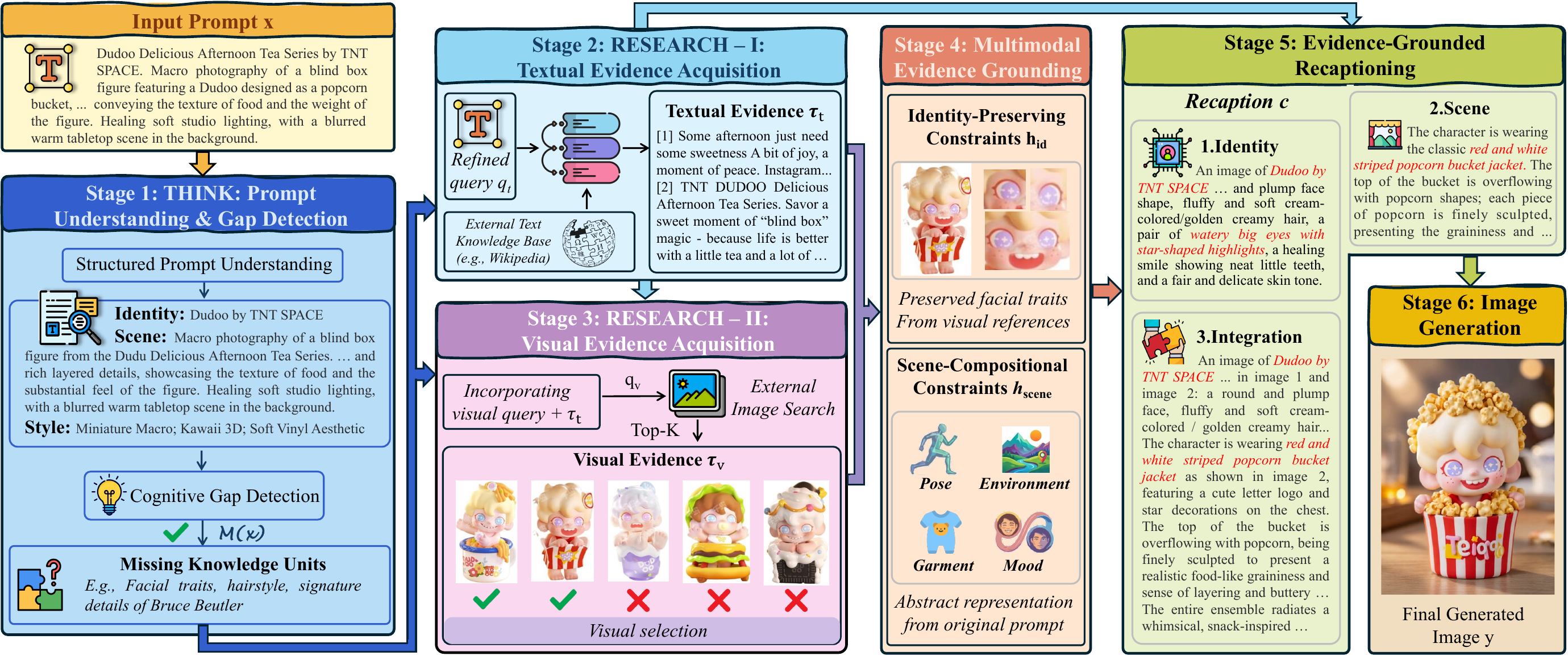}
    \vspace{-15pt}
    \caption{\textbf{Overview of the agentic pipeline of our method.}
Given an input prompt, our framework first performs prompt understanding and cognitive gap detection to identify missing but visually critical attributes. It then acquires complementary multimodal evidence through textual evidence searching and visual evidence searching. Based on the collected evidence, the model grounds the generation process with two types of constraints: identity-preserving constraints that capture character-specific visual traits, and scene-compositional constraints that specify pose, environment, garment, and overall mood. These grounded constraints are then integrated into an evidence-grounded recaptioning module, which produces a detailed caption for the downstream image generator to synthesize the final image.}
    \label{fig:method_overview}
    \vspace{-20pt}
\end{figure}

\subsection{Unified Fine-Tuning on Multimodal Agent Trajectories}
\label{sec:sft}

Building upon the unified multimodal formulation of Bagel described in Section~\ref{pril:bagel}, 
we further adapt the model to our agentic setting through supervised fine-tuning on interleaved multimodal trajectories from Section~\ref{sec:traing_data_curation}. 
Each training sample may contain textual reasoning traces, tool-use actions, recaptioning outputs, and image-generation targets, 
allowing the model to learn world-grounded generation as a unified sequential decision process. To facilitate efficient and unified training, we adopt a sequence packing strategy to concatenate heterogeneous samples into packed sequences, enabling the joint optimization of text-only and image-generation objectives within a single forward pass.

Our fine-tuning objective follows the ``think-to-generate'' paradigm of Bagel, employing a dual-loss design. Specifically, it combines a language modeling loss $\mathcal{L}_{\text{text}}$ for multimodal understanding and a latent-space regression loss $\mathcal{L}_{\text{image}}$ for image generation:
\begin{equation}
\mathcal{L}_{\text{SFT}} = \mathcal{L}_{\text{text}} + \mathcal{L}_{\text{image}}
\end{equation}

\paragraph{Text supervision.}
For tokens corresponding to reasoning, tool invocation, and recaptioning targets, 
we apply the autoregressive next-token prediction objective introduced in Section~\ref{pril:bagel}. 
In practice, this branch is instantiated as a token-level cross-entropy loss over supervision positions. 
To better preserve critical structural formats in agent trajectories, 
we further upweight special tokens associated with reasoning and action boundaries, 
such as \texttt{<think>}, \texttt{<tool\_call>}, and \texttt{<recaption>} tags. 
This encourages reliable learning of structured reasoning traces and executable output formats.

\paragraph{Image supervision.}
For samples involving image generation, we retain the latent flow-matching training objective from Section~\ref{pril:bagel}. 
Given a target image, we first encode it into a clean latent representation with the frozen VAE, 
then perturb the latent with Gaussian noise at a sampled timestep, 
and train the model to predict the corresponding flow velocity field in latent space. 
This branch enables the model to align generated images with both the input instruction and the retrieved multimodal evidence, 
while remaining compatible with the unified autoregressive backbone.

Importantly, the two objectives are activated by separate supervision masks within each packed sequence. 
In addition, we employ a hybrid attention masking strategy to regulate information flow across interleaved reasoning traces, retrieved reference images, recaption tokens, and generation tokens. 
This design is particularly important for agentic multimodal training, as it preserves the sequential structure of textual reasoning while preventing noisy historical traces from interfering with the final image synthesis stage. 
Detailed masking rules and visualizations are provided in Appendix~\ref{appendix:atten_mask}. 
As a result, text-only trajectory segments contribute only to $\mathcal{L}_{\text{text}}$, whereas image-generation segments contribute to both $\mathcal{L}_{\text{text}}$ and $\mathcal{L}_{\text{image}}$.

\subsection{Unify-Agent: Inference-Time Multimodal Agentic Pipeline}
\label{subsec:inference_pipeline}

As illustrated in Figure~\ref{fig:method_overview}, Unify-Agent formulates world-grounded image synthesis as a sequential inference-time decision process, rather than a single-shot mapping from prompt to image. 
This design directly instantiates the factorization in Eq.~\ref{eq:joint_prob}: given a user prompt $x$, the model first determines whether the request contains visually critical knowledge missing from its parametric memory, then actively acquires complementary multimodal evidence, converts the retrieved evidence into generation-relevant constraints, and finally synthesizes the image through an evidence-grounded recaption. 
By interleaving reasoning, searching, and synthesis within a unified execution policy, the agent is able to compensate for world-knowledge deficiencies while preserving the compositional intent of the original prompt.

\paragraph{\textsc{Think}: Prompt understanding and cognitive gap detection.}
Unify-Agent begins by interpreting the input prompt $x$ not as a self-sufficient description, but as a partially specified generation request whose visually critical details may be incomplete. 
To make this explicit, the model first performs structured prompt understanding, decomposing the instruction into semantically meaningful factors such as target identity, scene configuration, stylistic intent, and photographic requirements. 
Based on this structured representation, the model then estimates whether faithful synthesis depends on knowledge that is absent, ambiguous, or unreliable in its internal parametric memory. 
This process is captured by the latent gap variable
\begin{equation}
g \sim p_{\theta}(g \mid x),
\end{equation}
which governs whether external world knowledge should be invoked.

To further characterize the content of this gap, we represent the missing information as a set of knowledge units
\begin{equation}
\mathcal{M}(x) = \{m_1, \dots, m_K\},
\end{equation}
where each $m_k$ denotes an identity-critical or visually consequential attribute that is not sufficiently specified by the prompt itself, such as facial structure, hairstyle, signature appearance, object-specific structure, or scene-specific factual details. 
When $\mathcal{M}(x)\neq \emptyset$, the model enters the subsequent research stage to actively resolve these missing variables; otherwise, it may proceed directly to generation. 
Crucially, the purpose of this stage is not merely to recognize that an entity is rare or knowledge-intensive, but to determine which missing attributes must be recovered in order to support faithful world-grounded synthesis.

\paragraph{\textsc{Research}: Sequential multimodal evidence acquisition.}
Conditioned on the inferred gap state $g$, the agent proceeds to external evidence acquisition, corresponding to the factor
\begin{equation}
p_{\theta}(\tau_t, \tau_v \mid x, g).
\end{equation}
Rather than searching for all evidence in a single step, Unify-Agent adopts a sequential multimodal strategy in which textual evidence is acquired before visual evidence. 
This ordering is intentional: textual search first provides semantic grounding, identity disambiguation, and high-level background knowledge, which then guides a more precise visual search.

Concretely, the agent first formulates a textual query and obtains a textual evidence trace $\tau_t$, which provides compact semantic context about the target concept. 
This textual scaffold is then used to refine the subsequent visual search process, enabling the model to formulate a more informed visual query and retrieve a visual evidence trace $\tau_v$ composed of identity-relevant and context-compatible reference images. 
Compared with naive name-based retrieval, this sequential design allows evidence acquisition to be conditioned jointly on prompt intent and semantic grounding, thereby improving both retrieval precision and downstream grounding quality.

\paragraph{\textsc{Recaption}: Multimodal grounding into executable specifications.}
A central design choice in Unify-Agent is that retrieved evidence is not passed directly to the image generator in its raw form. 
As shown in Section~\ref{sec:motivation}, although naive knowledge injection is helpful, it remains suboptimal: raw text may introduce redundant or weakly visual information, while raw reference images alone may over-constrain the synthesis process. 
To address this mismatch, the agent first transforms the collected multimodal evidence into a set of grounded constraints that are directly useful for generation.

These constraints take two complementary forms. 
The first is an \emph{identity-preserving constraint}, which captures the visual attributes that must remain faithful to the target identity, such as facial structure, appearance cues, or object-defining characteristics. 
The second is a \emph{scene-compositional constraint}, which encodes the prompt-specified factors of generation, including pose, environment, garment, mood, composition, and overall presentation. 
Together, these grounded constraints serve as an abstraction layer that filters noisy raw evidence and reorganizes it into a more generation-compatible representation. 
The agent then integrates these constraints into an evidence-grounded recaption
\begin{equation}
c \sim p_{\theta}(c \mid x, g, \tau_t, \tau_v),
\end{equation}
which acts as the final executable specification for synthesis. 
In this way, recaptioning provides a structured textual interface that unifies the original user intent with retrieved semantic and visual evidence, while preserving both identity fidelity and compositional controllability.

\paragraph{\textsc{Generate}: Evidence-grounded image synthesis.}
Given the evidence-grounded recaption $c$ and the selected visual anchors $\tau_v$, the downstream generator finally synthesizes the image according to
\begin{equation}
y \sim p_{\theta}(y \mid c, \tau_v).
\end{equation}
This formulation reflects our modeling assumption that, once the external evidence has been distilled into a structured recaption and a compact set of visual anchors, the final synthesis process no longer needs to depend directly on the full upstream reasoning history $(x, g, \tau_t)$. 
Instead, generation is driven by refined, visually grounded constraints that are specifically tailored for the image generator.

This design brings two advantages. 
First, it prevents noisy reasoning traces or redundant textual evidence from interfering with the continuous generative process. 
Second, it preserves the complementary roles of the two conditioning signals: the recaption provides a unified and controllable description of the intended image, while the visual anchors stabilize identity realization and factual appearance. 
As a result, the generated output remains faithful to both the retrieved world knowledge and the compositional intent of the original prompt.


Overall, Unify-Agent reframes world-grounded image synthesis as an active agentic pipeline of \textsc{Think}, \textsc{Research}, \textsc{Recaption}, and \textsc{Generate}. 
Rather than treating the user prompt as a complete specification, the model identifies missing knowledge, actively acquires external evidence, reorganizes that evidence into a synthesis-ready representation, and finally produces an image grounded in both world knowledge and user intent. 
This unified pipeline moves beyond prompt-only generation and enables reliable synthesis of rare, long-tail, and knowledge-intensive concepts.

\section{Experiments}

\subsection{Experimental Details}

\paragraph{Baselines} We benchmark our Unify-Agent against three categories of baselines. First, we include leading commercial models such as Seedream-Family\citep{seedream2025seedream40nextgenerationmultimodal}, Nano Banana-2\citep{nanobanana2}, DALLE-3 \citep{betker2023improving}, and GPT-Image-1.5\citep{GPT1.5}, which serve as upper-bound references for factual text-to-image generation. Second, we consider representative generation-only models, including FLUX.1-dev \citep{flux1dev2025}, SD-3.5-large \citep{esser2024scaling}, Playground-v2.5\citep{playground2.5}, Z-Image\citep{team2025zimage} and Qwen-Image \citep{wu2025qwenimagetechnicalreport}, which are widely adopted open-source diffusion alternatives. Third, we evaluate competitive unified MLLMs, such as Janus-Pro-7B \citep{chen2025januspro}, Emu3.5\citep{cui2025emu35}, Echo-4o\citep{ye2025echo4oharnessingpowergpt4o} ,Hunyuan-Image-3.0\citep{hunyuanimage30}, and Bagel \citep{deng2025bagel}. These represent structurally comparable open-source baselines that natively unify visual understanding and generation. Most of the unified baselines share similar autoregressive or flow-matching training setups, ensuring comparability in architecture. Collectively, this suite provides a comprehensive coverage of both proprietary and open-source models across multiple paradigms.

\paragraph{Evaluation Setup}

Recently, using multimodal large language models (MLLMs) as automatic judges has become an increasingly common practice for multimodal evaluation, as it provides a scalable and efficient alternative to human annotation while enabling fine-grained assessment of multimodal outputs. Prior work on \textit{MLLM-as-a-Judge} \textbf{\citep{chen2024mllm}} suggests that MLLM-based evaluators can serve as practical proxies for expert judgment in vision-language tasks, particularly in scoring and pairwise comparison. Following this evaluation paradigm, we adopt MLLM-based evaluation on WISE \citep{niu2025wise}, T2I-FactualBench \citep{huang2025t2ifactualbench}, and Kitten \citep{huang2024kitten}, where ChatGPT-4o \citep{openai2024gpt4ocard} is used as the expert evaluator which helps maintain consistency with prior benchmark protocols. For our proposed FactIP Bench, we use Seed2.0 \citep{seedseed2} as the expert evaluator, considering its stronger multimodal judging capability. Additional details on the evaluation protocol, prompts, and scoring criteria are provided in the Appendix \ref{appen:eval_protocols}.

\subsection{Main Results}

\paragraph{FactIP} Tab. \ref{tab:factip-total} presents the quantitative results on our FactIP benchmark, which evaluates the capability of models to generate culturally significant concepts and intellectual properties with high factual fidelity. The evaluation is categorized into Character, Object, and Scene, assessing Clarity, Content, Aesthetics, and Relevance. Within the Unified MLLM category, our Unify-Agent achieves the highest Overall score of 73.2, significantly outperforming competitive proprietary and open-weight unified models. Notably, Unify-Agent exhibits exceptional performance in the critical Relevance dimension, securing the top scores across Character (67.3), Object (71.8), and Scene (78.2) sub-categories. This demonstrates its superior ability to preserve faithful IP identities and factual visual traits. Furthermore, Unify-Agent vastly surpasses traditional generation-only architectures like FLUX.1-dev (28.9) and SD-3.5-large (27.5), validating the effectiveness of integrating explicit multimodal reasoning and external knowledge for complex factual generation.

\begin{table*}[t]
    \caption{
    FactIP benchmark results with category-level weighted averages.
    \textbf{Character} includes celebrity, animation, game, comic, mythology, and mascot prompts;
    \textbf{Object} includes animals/plants, food, cultural relic/art, and toy prompts;
    \textbf{Scene} includes landmark and celebration prompts.
    \colorbox{purple!30}{\phantom{xx}} denotes the best score and \colorbox{purple!15}{\phantom{xx}} the second-best within each group.
    }
    \centering
    \footnotesize
    \setlength{\tabcolsep}{4pt}
    \resizebox{\textwidth}{!}{
    \begin{tabular}{c ccc ccc ccc ccc c}
    \toprule
    \multirow{2}{*}{\centering \textbf{Model}}
    & \multicolumn{3}{c}{\textbf{Clarity}}
    & \multicolumn{3}{c}{\textbf{Content}}
    & \multicolumn{3}{c}{\textbf{Aesthetics}}
    & \multicolumn{3}{c}{\textbf{Relevance}}
    & \multirow{2}{*}{\centering \textbf{Overall}} \\
    \cmidrule(lr){2-4} \cmidrule(lr){5-7} \cmidrule(lr){8-10} \cmidrule(lr){11-13}
    & Character & Object & Scene
    & Character & Object & Scene
    & Character & Object & Scene
    & Character & Object & Scene \\
    \midrule
    \rowcolor{blue!5}
    \multicolumn{14}{c}{\textbf{Commercial Models}} \\
    GPT-Image-1 & 94.6 & \cellcolor{purple!15}{94.1} & \cellcolor{purple!30}\textbf{96.2} & 83.5 & 71.1 & 78.4 & 89.2 & \cellcolor{purple!15}{88.1} & \cellcolor{purple!15}{94.1} & 61.9 & 60.2 & 76.4 & 69.3 \\
    GPT-Image-1.5 & 94.4 & \cellcolor{purple!30}\textbf{94.2} & \cellcolor{purple!15}{95.6} & 82.7 & 73.2 & 77.9 & 89.5 & \cellcolor{purple!15}{88.1} & 93.8 & 62.6 & 61.5 & 77.2 & 69.9 \\
    Seedream-4 & 93.9 & 83.7 & 88.0 & 81.0 & 92.3 & 78.1 & 88.2 & 81.7 & 91.5 & 80.5 & 90.5 & 86.0 & 83.0 \\
    Seedream-4.5 & 94.1 & 83.1 & 87.8 & 80.1 & 92.2 & 76.9 & 88.4 & 79.7 & 90.8 & 81.0 & 91.3 & 83.7 & 82.0 \\
    Seedream-5 & \cellcolor{purple!15}{95.0} & 86.0 & 88.1 & \cellcolor{purple!30}\textbf{86.7} & 94.3 & 83.5 & \cellcolor{purple!15}{90.3} & 87.1 & 91.7 & \cellcolor{purple!15}{84.3} & \cellcolor{purple!15}{91.8} & \cellcolor{purple!15}{87.1} & \cellcolor{purple!15}{87.3} \\
    Qwen-Image-2.0-Pro & 94.2 & 92.7 & 90.8 & 77.0 & 73.6 & 76.2 & 86.2 & 87.5 & 90.0 & 62.6 & 74.0 & 79.7 & 71.1 \\
    Nano Banana & 94.8 & 71.8 & 76.7 & 44.3 & \cellcolor{purple!30}\textbf{97.8} & \cellcolor{purple!15}{84.0} & 76.0 & 64.8 & 93.7 & 74.7 & 73.7 & 52.4 & 56.6 \\
    Nano Banana-Pro & 94.3 & 76.9 & 86.8 & 59.1 & 92.6 & 69.1 & 86.7 & 59.9 & 91.9 & 76.9 & 90.3 & 72.6 & 66.7 \\
    Nano Banana-2 & \cellcolor{purple!30}\textbf{96.6} & 88.2 & 90.4 & \cellcolor{purple!15}{86.3} & \cellcolor{purple!15}{96.0} & \cellcolor{purple!30}\textbf{86.6} & \cellcolor{purple!30}\textbf{92.2} & \cellcolor{purple!30}\textbf{92.2} & \cellcolor{purple!30}\textbf{95.3} & \cellcolor{purple!30}\textbf{85.5} & \cellcolor{purple!30}\textbf{93.9} & \cellcolor{purple!30}\textbf{89.5} & \cellcolor{purple!30}\textbf{88.5} \\
    \rowcolor{blue!5}
    \multicolumn{14}{c}{\textbf{Generation Only}} \\
    Pixel-Art-XL & 34.6 & 16.9 & 15.6 & 14.7 & 5.7 & 9.2 & 28.2 & 14.7 & 19.0 & 10.8 & 3.7 & 9.2 & 12.1 \\
    FLUX.1-schnell & 85.5 & 80.6 & 82.1 & 35.1 & 19.9 & 27.6 & 68.4 & 65.5 & 69.7 & 13.9 & 7.9 & 16.9 & 24.0 \\
    FLUX.1-dev & 92.5 & 89.0 & \cellcolor{purple!15}{91.8} & 51.9 & 40.4 & 37.7 & 78.4 & 75.8 & 80.3 & 17.0 & 8.2 & 18.2 & 28.9 \\
    FLUX.2-dev & 92.4 & \cellcolor{purple!15}{91.7} & 90.6 & \cellcolor{purple!30}\textbf{72.9} & \cellcolor{purple!15}{58.6} & 66.8 & \cellcolor{purple!15}{84.4} & \cellcolor{purple!30}\textbf{83.8} & \cellcolor{purple!15}{88.5} & \cellcolor{purple!30}\textbf{46.6} & \cellcolor{purple!30}\textbf{46.3} & \cellcolor{purple!15}{61.0} & \cellcolor{purple!30}\textbf{56.3} \\
    SD-3-medium & 85.1 & 81.7 & 78.5 & 40.9 & 22.0 & 28.6 & 70.6 & 64.9 & 66.8 & 16.2 & 6.6 & 17.9 & 25.7 \\
    SDXL-base-0.9 & 82.9 & 78.3 & 80.2 & 40.4 & 16.6 & 25.3 & 74.1 & 67.0 & 73.2 & 17.9 & 6.4 & 17.1 & 26.5 \\
    SD-3.5-large & 85.4 & 82.5 & 81.1 & 44.3 & 23.2 & 32.1 & 75.3 & 68.1 & 74.5 & 18.0 & 5.9 & 20.2 & 27.5 \\
    SD-3.5-medium & 84.3 & 73.6 & 79.2 & 40.9 & 17.6 & 31.6 & 72.5 & 54.6 & 67.4 & 17.7 & 7.1 & 20.1 & 26.5 \\
    Z-Image & \cellcolor{purple!15}{94.3} & \cellcolor{purple!30}\textbf{92.4} & \cellcolor{purple!30}\textbf{91.9} & \cellcolor{purple!15}{72.5} & \cellcolor{purple!30}\textbf{62.1} & \cellcolor{purple!15}{67.7} & \cellcolor{purple!30}\textbf{84.9} & \cellcolor{purple!15}{83.7} & 87.3 & 42.8 & \cellcolor{purple!15}{44.5} & 60.5 & 54.2 \\
    Qwen-Image & \cellcolor{purple!30}\textbf{94.5} & \cellcolor{purple!15}{91.7} & 91.7 & \cellcolor{purple!30}\textbf{72.9} & 57.3 & \cellcolor{purple!30}\textbf{67.9} & \cellcolor{purple!30}\textbf{84.9} & \cellcolor{purple!30}\textbf{83.8} & \cellcolor{purple!30}\textbf{89.4} & \cellcolor{purple!15}{45.3} & 43.1 & \cellcolor{purple!30}\textbf{62.6} & \cellcolor{purple!15}{55.4} \\
    \midrule
    \rowcolor{blue!5}
    \multicolumn{14}{c}{\textbf{Unified MLLM}} \\
    Janus-1.3B & 42.7 & 24.9 & 29.7 & 26.1 & 13.3 & 19.2 & 41.8 & 26.5 & 37.1 & 15.2 & 10.3 & 21.0 & 19.3 \\
    Janus-Pro & 60.0 & 40.3 & 43.9 & 40.1 & 25.6 & 28.5 & 58.2 & 46.4 & 56.6 & 24.3 & 21.4 & 30.2 & 30.3 \\
    Emu3 & 82.7 & 81.2 & 78.6 & 48.4 & 29.6 & 34.8 & 73.4 & 69.8 & 73.7 & 19.2 & 13.8 & 26.1 & 30.0 \\
    Echo4o & 89.9 & 90.7 & 89.7 & 65.1 & 49.3 & 61.0 & 80.8 & 80.5 & 85.0 & 36.1 & 35.7 & 52.0 & 47.3 \\
    Hunyuan-Image-3.0 & \cellcolor{purple!15}{93.0} & 91.1 & 90.0 & \cellcolor{purple!15}{72.3} & \cellcolor{purple!15}{59.7} & \cellcolor{purple!15}{70.4} & \cellcolor{purple!15}{84.6} & 83.0 & \cellcolor{purple!15}{87.7} & 40.6 & \cellcolor{purple!15}{46.8} & \cellcolor{purple!15}{63.3} & 53.4 \\
    Emu3.5 & \cellcolor{purple!30}\textbf{94.8} & \cellcolor{purple!30}\textbf{92.7} & \cellcolor{purple!30}\textbf{94.3} & 64.0 & 54.5 & 69.3 & \cellcolor{purple!30}\textbf{85.2} & \cellcolor{purple!15}{84.5} & \cellcolor{purple!30}\textbf{90.6} & \cellcolor{purple!15}{49.7} & 44.4 & 60.8 & \cellcolor{purple!15}{57.2} \\
    Bagel & 91.4 & \cellcolor{purple!15}{91.8} & 90.6 & 68.5 & 59.1 & 65.0 & 81.6 & 83.3 & 87.2 & 39.9 & 44.0 & 50.7 & 50.9 \\
    Bagel-CoT & 90.7 & 90.8 & 90.5 & 66.4 & 53.7 & 62.3 & 80.3 & 79.3 & 85.0 & 36.2 & 34.7 & 47.4 & 47.0 \\
    \textbf{Unify-Agent (Ours)} & 92.4 & 91.5 & \cellcolor{purple!15}{90.7} & \cellcolor{purple!30}\textbf{75.8} & \cellcolor{purple!30}\textbf{76.1} & \cellcolor{purple!30}\textbf{73.6} & 83.3 & \cellcolor{purple!30}\textbf{86.0} & 86.4 & \cellcolor{purple!30}\textbf{67.3} & \cellcolor{purple!30}\textbf{71.8} & \cellcolor{purple!30}\textbf{78.2} & \cellcolor{purple!30}\textbf{73.2} \\
    \bottomrule
    \end{tabular}
    }
    \label{tab:factip-total}
    \end{table*}

\paragraph{WiSE} Tab. \ref{tab:wise} reports results on the WiSE benchmark, which evaluates text-to-image generation performance across different knowledge dimensions, including cultural, time, space, biology, physics, and chemistry. Our Unify-Agent model attains the best Overall score of 0.77 within the Unified MLLM category, exceeding strong baselines such as BAGEL+CoT (0.70). Unify-Agent delivers the strongest performance among unified models across most individual domains, particularly excelling in cultural (0.82), biological (0.72) and chemistry (0.70) knowledge. These results indicate that Unify-Agent effectively integrates complex world knowledge into the visual generation process, narrowing the performance gap with commercial models.

\begin{table*}[t]
\caption{
Performance comparison on the WiSE benchmark across different knowledge dimensions (WiScore).
\colorbox{purple!30}{\phantom{xx}} denotes the best score and \colorbox{purple!15}{\phantom{xx}} the second-best within each group.
}
\centering
\scriptsize
\setlength{\tabcolsep}{9pt}
\resizebox{\textwidth}{!}{
\begin{tabular}{c cccccc c}
\toprule
\textbf{Model} & \textbf{Cultural} & \textbf{Time} & \textbf{Space} & \textbf{Biology} & \textbf{Physics} & \textbf{Chemistry} & \textbf{Overall} \\
\midrule
\rowcolor{blue!5}
\multicolumn{8}{c}{\textbf{Commercial Models}} \\
GPT-Image-1 & \cellcolor{purple!15}{0.81} & 0.71 & \cellcolor{purple!15}{0.89} & 0.83 & 0.79 & 0.74 & 0.80 \\
Nano Banana-Pro & \cellcolor{purple!30}\textbf{0.89} & \cellcolor{purple!15}{0.80} & \cellcolor{purple!15}{0.89} & \cellcolor{purple!15}{0.88} & \cellcolor{purple!15}{0.86} & \cellcolor{purple!30}\textbf{0.85} & \cellcolor{purple!15}{0.87} \\
Nano Banana & \cellcolor{purple!30}\textbf{0.89} & \cellcolor{purple!30}\textbf{0.87} & \cellcolor{purple!30}\textbf{0.95} & \cellcolor{purple!30}\textbf{0.89} & \cellcolor{purple!30}\textbf{0.89} & \cellcolor{purple!15}{0.79} & \cellcolor{purple!30}\textbf{0.89} \\
\midrule
\rowcolor{blue!5}
\multicolumn{8}{c}{\textbf{Generation Only}} \\
SD-v1-5 & 0.34 & 0.35 & 0.32 & 0.28 & 0.29 & 0.21 & 0.32 \\
SD-2-1 & 0.30 & 0.38 & 0.35 & 0.33 & 0.34 & 0.21 & 0.32 \\
FLUX.1-schnell & 0.39 & 0.44 & 0.50 & 0.31 & 0.44 & 0.26 & 0.40 \\
SD-3-medium & 0.42 & 0.44 & 0.48 & 0.39 & 0.47 & 0.29 & 0.42 \\
SD-XL-base-0.9 & 0.43 & 0.48 & 0.47 & 0.44 & 0.45 & 0.27 & 0.43 \\
SD-3.5-medium & 0.43 & 0.50 & 0.52 & 0.41 & 0.53 & 0.33 & 0.45 \\
SD-3.5-large & 0.44 & 0.50 & 0.58 & 0.44 & 0.52 & 0.31 & 0.46 \\
PixArt-Alpha & 0.45 & 0.50 & 0.48 & \cellcolor{purple!15}{0.49} & \cellcolor{purple!15}{0.56} & 0.34 & 0.47 \\
playground-v2.5 & \cellcolor{purple!15}{0.49} & \cellcolor{purple!15}{0.58} & 0.55 & 0.43 & 0.48 & 0.33 & 0.49 \\
FLUX.1-dev & 0.48 & \cellcolor{purple!15}{0.58} & \cellcolor{purple!15}{0.62} & 0.42 & 0.51 & \cellcolor{purple!15}{0.35} & \cellcolor{purple!15}{0.50} \\
Qwen-Image & \cellcolor{purple!30}\textbf{0.62} & \cellcolor{purple!30}\textbf{0.63} & \cellcolor{purple!30}\textbf{0.77} & \cellcolor{purple!30}\textbf{0.57} & \cellcolor{purple!30}\textbf{0.75} & \cellcolor{purple!30}\textbf{0.40} & \cellcolor{purple!30}\textbf{0.62} \\
\midrule
\rowcolor{blue!5}
\multicolumn{8}{c}{\textbf{Unified MLLM}} \\
Janus-1.3B & 0.16 & 0.26 & 0.35 & 0.28 & 0.30 & 0.14 & 0.23 \\
Janus-Pro-1B & 0.20 & 0.28 & 0.45 & 0.24 & 0.32 & 0.16 & 0.26 \\
vila-u-7b-256 & 0.26 & 0.33 & 0.37 & 0.35 & 0.39 & 0.23 & 0.31 \\
Janus-Pro-7B & 0.30 & 0.37 & 0.49 & 0.36 & 0.42 & 0.26 & 0.35 \\
Emu3 & 0.34 & 0.45 & 0.48 & 0.41 & 0.45 & 0.27 & 0.39 \\
Hunyuan-Image-3.0  & 0.58 & 0.57 & 0.70 & 0.56 & 0.63 & 0.31 & 0.57 \\
BLIP3o-8B & 0.49 & 0.51 & 0.63 & 0.54 & 0.63 & 0.37 & 0.52 \\
MetaQuery-XL & 0.56 & 0.55 & 0.62 & 0.49 & 0.63 & 0.41 & 0.55 \\
UniWorld-V1 & 0.53 & 0.55 & 0.73 & 0.45 & 0.59 & 0.41 & 0.55 \\
BAGEL & 0.44 & 0.55 & 0.68 & 0.44 & 0.60 & 0.39 & 0.52 \\
BAGEL+CoT & \cellcolor{purple!15}{0.76} & \cellcolor{purple!15}{0.69} & \cellcolor{purple!30}\textbf{0.75} & \cellcolor{purple!15}{0.65} & \cellcolor{purple!30}\textbf{0.75} & \cellcolor{purple!15}{0.58} & \cellcolor{purple!15}{0.70} \\
\textbf{Unify-Agent (Ours)} & \cellcolor{purple!30}\textbf{0.82} & \cellcolor{purple!30}\textbf{0.75} & \cellcolor{purple!15}{0.74} & \cellcolor{purple!30}\textbf{0.72} & \cellcolor{purple!15}{0.73} & \cellcolor{purple!30}\textbf{0.70} & \cellcolor{purple!30}\textbf{0.77} \\
\bottomrule
\end{tabular}
}
\label{tab:wise}
\vspace{-10pt}
\end{table*}

\paragraph{KiTTEN} Tab. \ref{tab:kitten} summarizes the quantitative results on the KiTTEN benchmark, which evaluates fine-grained generation capabilities along two primary dimensions: text alignment and entity alignment. The evaluation spans eight diverse categories, such as Aircraft, Vehicle, and Cuisine. Our Unify-Agent establishes a new state of the art among both generation-only and unified MLLM approaches with an overall score of 4.08, outperforming the strong baseline Imagen-3 (3.50). Unify-Agent achieves the highest overall text alignment (4.22) and entity alignment (3.93) scores, maintaining balanced and superior performance across all sub-categories. This demonstrates the model's robust capability to preserve intricate visual details and faithfully adhere to specific entity constraints.

\begin{table*}[t]
\caption{
KiTTEN benchmark results (scores range from 0 to 5). Each category shows Text alignment and Entity alignment scores.
\colorbox{purple!30}{\phantom{xx}} denotes the best score and \colorbox{purple!15}{\phantom{xx}} the second-best within each group.
}
\centering
\footnotesize
\setlength{\tabcolsep}{4pt}
\resizebox{\textwidth}{!}{
\begin{tabular}{c cc cc cc cc cc cc cc cc ccc}
\toprule
\multirow{2}{*}{\centering \textbf{Model}}
& \multicolumn{2}{c}{\textbf{Aircraft}}
& \multicolumn{2}{c}{\textbf{Vehicle}}
& \multicolumn{2}{c}{\textbf{Cuisine}}
& \multicolumn{2}{c}{\textbf{Flower}}
& \multicolumn{2}{c}{\textbf{Insect}}
& \multicolumn{2}{c}{\textbf{Landmark}}
& \multicolumn{2}{c}{\textbf{Plant}}
& \multicolumn{2}{c}{\textbf{Sport}}
& \multicolumn{3}{c}{\textbf{Overall}} \\
\cmidrule(lr){2-3} \cmidrule(lr){4-5} \cmidrule(lr){6-7} \cmidrule(lr){8-9} \cmidrule(lr){10-11} \cmidrule(lr){12-13} \cmidrule(lr){14-15} \cmidrule(lr){16-17} \cmidrule(lr){18-20}
& Text & Entity & Text & Entity & Text & Entity & Text & Entity & Text & Entity & Text & Entity & Text & Entity & Text & Entity & Text & Entity & All \\
\midrule
\rowcolor{blue!5}
\multicolumn{20}{c}{\textbf{Generation Only}} \\
SD & 3.06 & 2.34 & 3.62 & 3.12 & 2.97 & 3.09 & \cellcolor{purple!15}{4.22} & 2.28 & 2.84 & 1.22 & 3.78 & 1.97 & 3.69 & 1.69 & 3.47 & \cellcolor{purple!15}{3.44} & 3.46 & 2.39 & 2.92 \\
Flux & \cellcolor{purple!15}{3.31} & 2.41 & \cellcolor{purple!30}\textbf{4.31} & 2.97 & \cellcolor{purple!30}\textbf{4.28} & 2.47 & 4.12 & 1.75 & \cellcolor{purple!15}{3.50} & 1.12 & \cellcolor{purple!30}\textbf{3.94} & 1.72 & \cellcolor{purple!15}{4.44} & 1.22 & \cellcolor{purple!15}{3.75} & 2.69 & \cellcolor{purple!15}{3.96} & 2.04 & 3.00 \\
Custom-Diff & 3.22 & 2.94 & 3.50 & 3.41 & 3.53 & 1.78 & 3.72 & \cellcolor{purple!15}{3.28} & 2.66 & 1.88 & 3.44 & 3.34 & 3.47 & \cellcolor{purple!15}{2.91} & 2.75 & 2.03 & 3.29 & 2.70 & 3.00 \\
Imagen & 3.16 & 2.81 & 3.78 & 3.16 & 3.75 & 3.25 & 4.12 & 2.72 & 2.88 & 1.12 & 3.53 & 2.41 & 3.94 & 1.59 & \cellcolor{purple!15}{3.75} & 3.19 & 3.61 & 2.53 & 3.07 \\
DreamBooth & 3.16 & 3.12 & 3.66 & \cellcolor{purple!15}{3.50} & 3.31 & 2.66 & 3.88 & 2.59 & 2.94 & \cellcolor{purple!15}{1.91} & 3.53 & \cellcolor{purple!15}{3.53} & 3.41 & 2.59 & 3.09 & 2.66 & 3.37 & 2.82 & 3.09 \\
Instruct-Imagen & 2.53 & \cellcolor{purple!30}\textbf{3.88} & 3.25 & \cellcolor{purple!30}\textbf{4.06} & 2.75 & \cellcolor{purple!15}{3.91} & 2.41 & \cellcolor{purple!30}\textbf{4.03} & 1.91 & \cellcolor{purple!30}\textbf{3.22} & 2.94 & \cellcolor{purple!30}\textbf{4.09} & 3.00 & \cellcolor{purple!30}\textbf{3.31} & 2.28 & 3.25 & 2.63 & \cellcolor{purple!30}\textbf{3.72} & \cellcolor{purple!15}{3.17} \\
Imagen-3 & \cellcolor{purple!30}\textbf{3.41} & \cellcolor{purple!15}{3.19} & \cellcolor{purple!15}{4.25} & 3.34 & \cellcolor{purple!15}{4.25} & \cellcolor{purple!30}\textbf{4.06} & \cellcolor{purple!30}\textbf{4.38} & 3.03 & \cellcolor{purple!30}\textbf{4.09} & 1.31 & \cellcolor{purple!15}{3.88} & 2.25 & \cellcolor{purple!30}\textbf{4.72} & 1.69 & \cellcolor{purple!30}\textbf{4.41} & \cellcolor{purple!30}\textbf{3.75} & \cellcolor{purple!30}\textbf{4.17} & \cellcolor{purple!15}{2.83} & \cellcolor{purple!30}\textbf{3.50} \\
\midrule
\rowcolor{blue!5}
\multicolumn{20}{c}{\textbf{Unified MLLM}} \\
Bagel & 3.51 & \cellcolor{purple!15}{2.97} & \cellcolor{purple!15}{4.46} & \cellcolor{purple!15}{4.08} & 3.91 & 3.40 & 3.44 & 2.64 & 2.65 & 1.58 & 3.31 & 1.79 & 3.01 & 1.71 & 3.23 & 2.98 & 3.44 & \cellcolor{purple!15}{2.64} & \cellcolor{purple!15}{3.04} \\
Bagel-CoT & \cellcolor{purple!15}{3.67} & 2.94 & 4.42 & 3.98 & \cellcolor{purple!15}{4.10} & \cellcolor{purple!15}{4.01} & \cellcolor{purple!15}{3.56} & \cellcolor{purple!15}{3.12} & \cellcolor{purple!15}{2.98} & \cellcolor{purple!15}{2.12} & \cellcolor{purple!15}{3.41} & \cellcolor{purple!15}{2.11} & \cellcolor{purple!15}{3.31} & \cellcolor{purple!15}{2.26} & \cellcolor{purple!15}{3.84} & \cellcolor{purple!15}{3.63} & \cellcolor{purple!15}{3.66} & 2.02 & 2.84 \\
\textbf{Unify-Agent (Ours)} & \cellcolor{purple!30}\textbf{4.03} & \cellcolor{purple!30}\textbf{3.59} & \cellcolor{purple!30}\textbf{4.54} & \cellcolor{purple!30}\textbf{4.09} & \cellcolor{purple!30}\textbf{4.36} & \cellcolor{purple!30}\textbf{4.13} & \cellcolor{purple!30}\textbf{4.28} & \cellcolor{purple!30}\textbf{4.16} & \cellcolor{purple!30}\textbf{4.17} & \cellcolor{purple!30}\textbf{3.97} & \cellcolor{purple!30}\textbf{3.99} & \cellcolor{purple!30}\textbf{3.76} & \cellcolor{purple!30}\textbf{4.46} & \cellcolor{purple!30}\textbf{3.81} & \cellcolor{purple!30}\textbf{3.95} & \cellcolor{purple!30}\textbf{3.92} & \cellcolor{purple!30}\textbf{4.22} & \cellcolor{purple!30}\textbf{3.93} & \cellcolor{purple!30}\textbf{4.08} \\
\bottomrule
\end{tabular}
}
\label{tab:kitten}
\vspace{-10pt}
\end{table*}

\paragraph{T2I-FactBench} Tab. \ref{tab:t2ifact} presents results on T2I-FactBench, a three-tiered benchmark measuring the factual accuracy of T2I models: Single Knowledge Concept Memorization (SKCM), Instantiation (SKCI), and Multiple Concept Composition with Interaction (MKCC). Within the unified MLLM group, Unify-Agent achieves top overall scores in SKCI (77.4) and MKCC (71.5). It also records an SKCM concept score of 69.2, comparing favorably against commercial models like DALLE-3 (55.5). This progression indicates that Unify-Agent captures the intrinsic visual attributes of single concepts (SKCM), adapts them to varied conditions like differing actions or scenes (SKCI), and handles the explicit semantic relationships necessary for multi-concept interactions (MKCC).

\begin{table*}[t]
\caption{
Comparison on T2I-FactBench (SKCM, SKCI, and MKCC).
\colorbox{purple!30}{\phantom{xx}} denotes the best score and \colorbox{purple!15}{\phantom{xx}} the second-best within each group.
}
\centering
\footnotesize
\setlength{\tabcolsep}{8pt}
\resizebox{\textwidth}{!}{
\begin{tabular}{c c ccc cccc}
\toprule
\multirow{2}{*}{\centering \textbf{Model}}
& \multicolumn{1}{c}{\textbf{SKCM}}
& \multicolumn{3}{c}{\textbf{SKCI}}
& \multicolumn{4}{c}{\textbf{MKCC}} \\
\cmidrule(lr){2-2} \cmidrule(lr){3-5} \cmidrule(lr){6-9}
& Concept & Concept & Instantiation & All & Concept & Instantiation & Composition & All \\
\midrule
\rowcolor{blue!5}
\multicolumn{9}{c}{\textbf{Generation Only}} \\
SD v1.5 & 40.5 & 52.9 & 53.9 & 53.4 & 37.6 & 13.4 & 15.1 & 22.0 \\
Pixart & 26.4 & 46.2 & 55.3 & 50.8 & 35.8 & 19.8 & 24.3 & 26.6 \\
SD XL & 45.8 & 59.9 & 65.8 & 62.8 & 51.7 & 28.0 & 35.4 & 38.4 \\
Playground & 45.6 & \cellcolor{purple!15}{66.1} & 62.5 & 64.3 & 53.8 & 35.4 & 44.8 & 44.7 \\
Flux.1-dev & 35.6 & 54.9 & 58.0 & 56.5 & 56.9 & 54.1 & 63.8 & 58.3 \\
SD-3.5 & \cellcolor{purple!15}{46.2} & 64.6 & \cellcolor{purple!15}{71.2} & \cellcolor{purple!15}{67.9} & \cellcolor{purple!15}{68.9} & \cellcolor{purple!15}{59.2} & \cellcolor{purple!15}{75.5} & \cellcolor{purple!15}{67.9} \\
DALLE-3 & \cellcolor{purple!30}\textbf{55.5} & \cellcolor{purple!30}\textbf{72.4} & \cellcolor{purple!30}\textbf{88.5} & \cellcolor{purple!30}\textbf{80.5} & \cellcolor{purple!30}\textbf{71.3} & \cellcolor{purple!30}\textbf{70.2} & \cellcolor{purple!30}\textbf{85.6} & \cellcolor{purple!30}\textbf{75.7} \\
\midrule
\rowcolor{blue!5}
\multicolumn{9}{c}{\textbf{Unified MLLM}} \\
Bagel & 31.8 & 52.3 & 64.2 & 58.2 & \cellcolor{purple!15}{68.1} & 57.6 & 73.2 & 66.3 \\
Bagel-CoT & \cellcolor{purple!15}{34.6} & \cellcolor{purple!15}{52.6} & \cellcolor{purple!15}{64.5} & \cellcolor{purple!15}{58.5} & 67.2 & \cellcolor{purple!15}{60.3} & \cellcolor{purple!30}\textbf{77.1} & \cellcolor{purple!15}{68.2} \\
\textbf{Unify-Agent (Ours)} & \cellcolor{purple!30}\textbf{69.2} & \cellcolor{purple!30}\textbf{75.3} & \cellcolor{purple!30}\textbf{79.6} & \cellcolor{purple!30}\textbf{77.4} & \cellcolor{purple!30}\textbf{76.1} & \cellcolor{purple!30}\textbf{64.8} & \cellcolor{purple!15}{73.6} & \cellcolor{purple!30}\textbf{71.5} \\
\bottomrule
\end{tabular}
}
\label{tab:t2ifact}
\vspace{-10pt}
\end{table*}

\begin{table}[t]
    \caption{
    Ablation study of Unify-Agent on the FactIP benchmark.
    We report top-level weighted averages over \textbf{Clarity}, \textbf{Content}, \textbf{Aesthetics}, and \textbf{Relevance}.
    The middle rows show expected performance drops when removing key pipeline components, grounded constraint types, or recaption-stage architectural modules.
    Numbers in parentheses denote changes relative to the vanilla Bagel baseline; red indicates improvement and green indicates degradation.
    The best result in each column is marked in \textbf{bold}, and the second-best is \underline{underlined}.
    }
    \centering
    \scriptsize
    \setlength{\tabcolsep}{8pt}
    \newcommand{\gain}[1]{\textcolor{red!80!black}{(#1)}}
    \newcommand{\drop}[1]{\textcolor{green!50!black}{(#1)}}
    \newcommand{\same}[1]{\textcolor{gray}{(#1)}}
    \resizebox{\columnwidth}{!}{
    \begin{tabular}{l ccccc}
    \toprule
    \textbf{Variant} & \textbf{Clarity} & \textbf{Content} & \textbf{Aesthetics} & \textbf{Relevance} & \textbf{Overall} \\
    \midrule
    \rowcolor{gray!10}
    Baseline (Vanilla Bagel) & 91.3  & 64.2  & 84.0  & 44.9  & 50.9  \\
    \midrule
    \rowcolor{blue!5}
    \multicolumn{6}{c}{\textbf{Pipeline Ablations}} \\
    w/o Text-Search & 90.7 \drop{-0.6} & 70.9 \gain{+6.7} & 84.3 \gain{+0.3} & 64.6 \gain{+19.7} & 65.4 \gain{+14.5} \\
    w/o Image-Search & \textbf{92.1} \gain{+0.8} & 73.1 \gain{+8.9} & \underline{85.0} \gain{+1.0} & 50.8 \gain{+5.9} & 56.2 \gain{+5.3} \\
    w/o Recaption & 83.0 \drop{-8.3} & 69.0 \gain{+4.8} & 74.5 \drop{-9.5} & 60.2 \gain{+15.3} & 62.9 \gain{+12.0} \\
    \midrule
    \rowcolor{blue!5}
    \multicolumn{6}{c}{\textbf{Constraint Ablations}} \\
    Recaption w/o Identity-preserving & \underline{91.5} \gain{+0.2} & 72.6 \gain{+8.4} & 83.4 \drop{-0.6} & 65.9 \gain{+21.0} & 67.7 \gain{+16.8} \\
    Recaption w/o Scene-compositional & 90.7 \drop{-0.6} & 70.8 \gain{+6.6} & 80.7 \drop{-3.3} & 68.6 \gain{+23.7} & 68.2 \gain{+17.3} \\
    \midrule
    \rowcolor{blue!5}
    \multicolumn{6}{c}{\textbf{Recaption Architecture Ablations}} \\
    Recaption w/o VAE & 90.9 \drop{-0.4} & \underline{74.3} \gain{+10.1} & 84.5 \gain{+0.5} & \underline{70.8} \gain{+25.9} & \underline{71.2} \gain{+20.3} \\
    Recaption w/o ViT & 88.6 \drop{-2.7} & 68.4 \gain{+4.2} & 81.1 \drop{-2.9} & 58.7 \gain{+13.8} & 61.4 \gain{+10.5} \\
    \midrule
    \rowcolor{purple!10}
    \textbf{Unify-Agent (Full)} & 91.2 \drop{-0.1} & \textbf{75.2} \gain{+11.0} & \textbf{85.2} \gain{+1.2} & \textbf{72.4} \gain{+27.5} & \textbf{73.2} \gain{+22.3} \\
    \bottomrule
    \end{tabular}
    }
    \label{tab:ablation}
\end{table}

\subsection{Analysis of Experimental Results}

\subsubsection{Ablation Study}

Tab.~\ref{tab:ablation} presents the ablation results of Unify-Agent on the FactIP benchmark. 
Compared with the vanilla Bagel baseline, the full model achieves the best performance on all metrics, improving the Overall score from 50.9 to 73.2. 
The largest gain is observed on \textbf{Relevance} (44.9 $\rightarrow$ 72.4), indicating that the main advantage of Unify-Agent lies in stronger factual grounding and identity fidelity, rather than merely improving generic image quality.

Pipeline-level ablations confirm the contribution of each component. 
Removing \emph{text search} reduces the Overall score to 65.4, showing that textual retrieval provides an important semantic scaffold for disambiguation and high-level factual grounding. 
Removing \emph{image search} causes a larger degradation, especially on Relevance (72.4 $\rightarrow$ 50.8), and lowers the Overall score to 56.2, which highlights the importance of visual evidence for preserving fine-grained identity and appearance details. 
Removing \emph{recaptioning} also leads to a substantial drop (Overall 62.9), verifying that raw retrieved evidence is not an optimal conditioning signal and must be reorganized into a compact, executable specification.

Constraint-level ablations show a similarly clear pattern. 
Without \emph{identity-preserving} constraints, Relevance drops to 65.9, confirming their role in maintaining subject-specific fidelity. 
Without \emph{scene-compositional} constraints, Content and Aesthetics decline to 70.8 and 80.7, respectively, indicating their importance for faithfully instantiating prompt-specified scene attributes. 
Overall, these results show that the gains of Unify-Agent arise from the joint effect of sequential multimodal evidence acquisition, grounded recaptioning, and structured generation constraints.

\begin{figure}[t]
    \centering
    \includegraphics[width=1.00\textwidth,keepaspectratio]{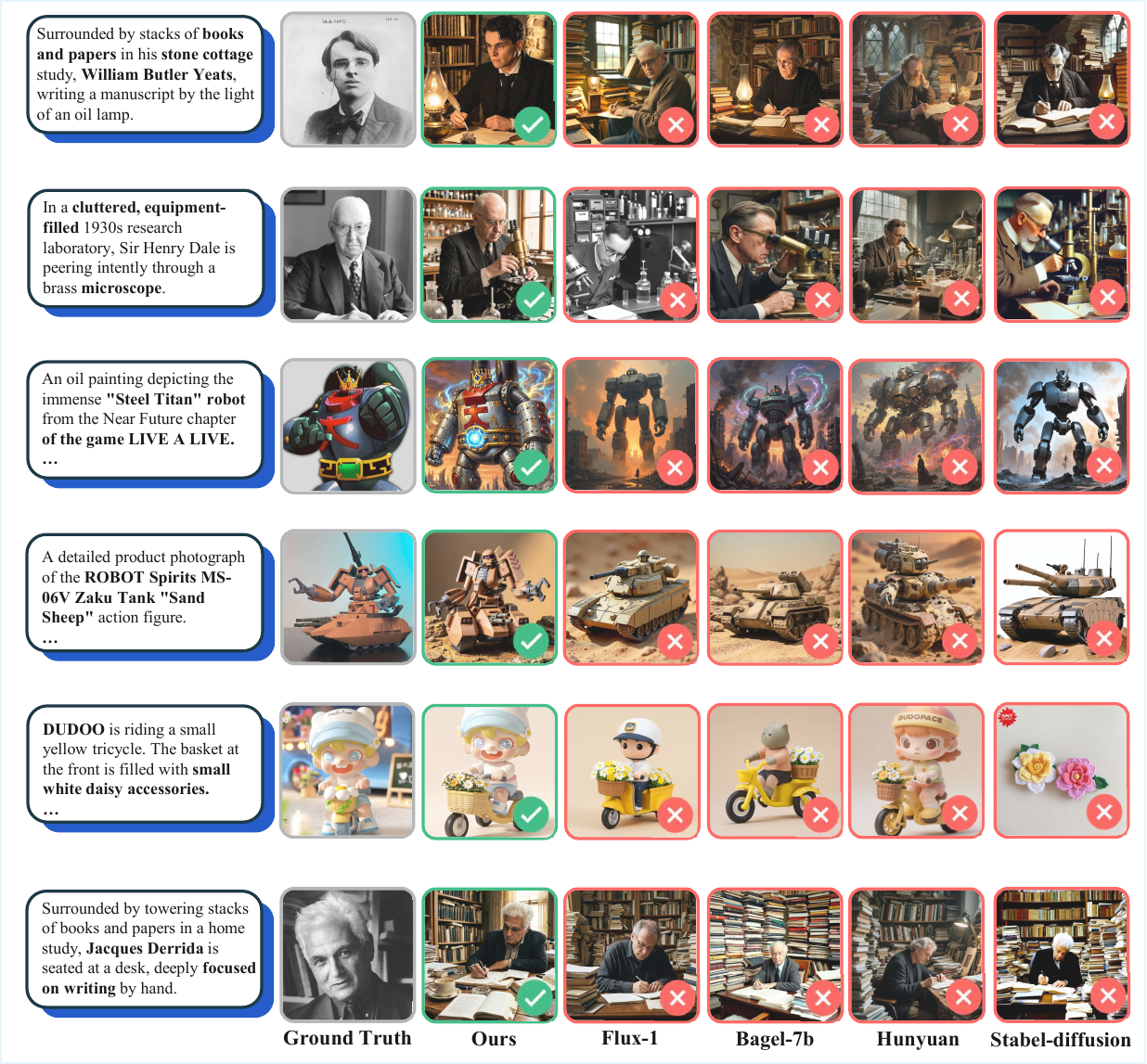}
    \caption{Qualitative comparison of multi-image generation results on knowledge-intensive prompts involving historical figures, fictional characters, products, and stylized toys. Our method consistently produces images that better preserve subject identity, fine-grained attributes, and prompt-specific details, while achieving stronger real-world knowledge grounding than competing baselines, including Flux-1, Bagel-7b, Hunyuan, and Stable Diffusion.}
    \label{fig:showcase}
    \vspace{-10pt}
\end{figure}

\subsection{Finding: Why Generation Helps Understanding in a Unified Model}
\label{subsec:gen_help_understanding}

The ablations in Tab.~\ref{tab:ablation} provide an additional insight beyond component-level contributions: in a unified understanding-generation model, generation can in fact improve understanding. 
This effect is particularly evident in the recaption stage, where the model must transform retrieved reference images into a structured textual specification that is both semantically faithful and generation-ready. 
In other words, recaption is not a pure language task, but a multimodal understanding problem that requires the model to identify \emph{what} appears in the image, \emph{which} attributes are identity-critical, and \emph{how} these visual cues should be expressed for downstream synthesis.

This observation helps explain why an end-to-end unified architecture is beneficial. 
In Bagel-style unified understanding-generation models, visual inputs are tokenized through a dual \textit{VAE}+\textit{ViT} design. Within this parallel architecture, the VAE extracts low-level visual latents that preserve appearance details such as color, texture, material, and local structure, while the ViT simultaneously encodes the input images into higher-level semantic tokens that are more suitable for visual understanding and recaptioning.
As ViT is effective at modeling global context and long-range dependencies, it functions as a semantic visual encoder that helps the language branch reason about object identity, attribute relations, and scene composition. 
At the same time, the VAE complements this process by retaining fine-grained perceptual priors that are often crucial for identity-sensitive grounding.

The recaption architecture ablations in Tab.~\ref{tab:ablation} are consistent with this interpretation. 
Removing ViT causes a substantial drop in performance, especially on Relevance and Overall, indicating that high-level semantic visual tokens are critical for accurately understanding the retrieved reference images. 
Removing VAE also degrades performance, albeit more moderately, suggesting that low-level appearance cues remain helpful even when higher-level semantic reasoning is available. 
Together, these results suggest that the benefit of a unified model does not come only from parameter sharing, but from the joint availability of low-level generative latents and high-level semantic visual representations within a single end-to-end framework. 
This synergy allows the model to produce better recaptions, which in turn leads to better grounded image synthesis.

\subsubsection{Visualization comparison results}

Figure~\ref{fig:showcase} shows that our \textbf{Unify-Agent} consistently outperforms Flux-1, Bagel-7b, Hunyuan, and Stable Diffusion on knowledge-intensive and attribute-sensitive prompts. Across diverse cases including historical figures, fictional entities, fine-grained products, and stylized toys, our method generates images with superior subject fidelity, more accurate attribute binding, and stronger prompt adherence. In particular, \textbf{Unify-Agent} better captures entity-specific identity cues and long-tail concepts, whereas competing methods often suffer from identity drift, missing attributes, or degeneration into generic substitutes. These qualitative results demonstrate that, by combining unified multi-image generation with agentic search for external world knowledge, \textbf{Unify-Agent} achieves more accurate, consistent, and knowledge-grounded visual generation.

\section{Conclusion}

We present \textbf{Unify-Agent}, a unified multimodal agent for world-grounded image synthesis. 
Rather than treating image generation as a closed-book prompt-to-image mapping, Unify-Agent formulates it as an inference-time process of cognitive gap detection, multimodal evidence acquisition, grounded recaptioning, and final synthesis. 
By explicitly resolving missing world knowledge through retrieved textual and visual evidence, our framework enables more faithful generation of rare, long-tail, and knowledge-intensive concepts. 
More broadly, our findings suggest that world-grounded image synthesis is a distinct multimodal reasoning problem that requires tightly coupled understanding, acting, and generation. 
We hope this work motivates future research on more reliable and open-world generative agents.

\section*{Limitations and Future Work}

Our work still has several limitations. Current open-source unified multimodal models remain substantially weaker than the strongest closed-source systems; for example, Bagel still has limited long-context capability and can support only a relatively small number of images within a single context, which constrains more complex agent behaviors. Moreover, although we demonstrate strong results on IP- and concept-centric world-grounded synthesis, our current pipeline is still limited to a relatively shallow one-pass workflow, rather than more general iterative agent behaviors such as interleaved text-image search, reflection, and replanning, which are crucial for harder open-world tasks such as travel planning or academic report generation. At the same time, our results provide encouraging evidence for the feasibility of end-to-end unified agent training and highlight the advantage of unified generation and reasoning multimodal models, where generation and understanding can mutually reinforce each other. In future work, we plan to validate these findings on stronger unified backbones and extend the framework toward more capable multimodal agents that can support longer-horizon planning, repeated search, and adaptive reasoning in more complex real-world settings.

\newpage

\bibliographystyle{plainnat}
\bibliography{ref}

\newpage

\appendix

\section*{Appendix}

\section{Implementation Details}
\label{Apendex:sec imple detail}
The Supervised Fine-Tuning (SFT) is conducted on our self-constructed interleaved image-and-text dataset with 64 NVIDIA H20 GPUs. The entire fine-tuning process lasted for approximately 10 days, completing a total of 10k gradient steps. For the training objective, our model is optimized using a combined loss formulation that jointly updates the language model, the ViT encoder, and the multimodal connectors, while keeping the continuous generation VAE weights strictly frozen. Specifically, we apply a standard Cross-Entropy (CE) loss for autoregressive text generation and visual token prediction, alongside a Mean Squared Error (MSE) loss on the image-reconstruction latents for the visual generation branch. Both the CE and MSE losses are equally scaled with a weight of 1.0. Furthermore, to improve the model's instruction-following capability in multimodal contexts, we enable CE loss reweighting, which specifically increases the penalty weight on designated special tokens. To stabilize the optimization over ultra-long interleaved sequences, we apply a constant learning rate scheduler with a peak learning rate of $5 \times 10^{-5}$ after 500 linear warmup steps, and enforce gradient clipping with a maximum L2 norm of 5.0. A comprehensive list of hyperparameters for the SFT stage is summarized in Table~\ref{tab:hyperparameters}.

\begin{table}[htbp]
\centering
\captionsetup{justification=centering}
\caption{Hyperparameters for the Supervised Fine-Tuning (SFT) stage.}
\label{tab:hyperparameters}
\begin{tabular}{@{}ll@{}}
\toprule
\textbf{Hyperparameter} & \textbf{Value} \\ \midrule
\multicolumn{2}{c}{\textit{Model \& Architecture Setup}} \\ \midrule
Base LLM & Bagel-14B \\
Vision Encoder (ViT) & SigLIP-SO400M-14 (NaViT) \\
Visual Generation VAE & FLUX VAE (Frozen) \\
ViT Patch Size & 14 \\
VAE Latent Patch Size & 2 \\
Tied Word Embeddings & False \\ \midrule
\multicolumn{2}{c}{\textit{Data \& Sequence Length}} \\ \midrule
Max Tokens per Sample & 40,240 \\
Expected Tokens per Batch & 40,240 \\
Max Packed Tokens (Hard Limit) & 41,520 \\
Vit Transform & (378, 980) \\
Vae Transform & (512, 1024) \\ \midrule
\multicolumn{2}{c}{\textit{Optimization \& Objective}} \\ \midrule
Optimizer & AdamW \\
Adam $(\beta_1, \beta_2, \epsilon)$ & $(0.9, 0.95, 10^{-15})$ \\
Learning Rate & $5 \times 10^{-5}$ \\
LR Scheduler & Constant \\
Warmup Steps & 500 \\
Total Training Steps & 10,000 \\
Max Gradient Norm & 5.0 \\
Cross-Entropy (CE) Weight & 1.0 \\
Reconstruction (MSE) Weight & 1.0 \\
CE Loss Special Token Reweighting & True \\
Special Token CE Weight & 3.0 \\\midrule
\multicolumn{2}{c}{\textit{Hardware \& Distributed Setup}} \\ \midrule
Hardware & $64 \times$ NVIDIA H20 GPUs \\
Training Duration & $\sim 10$ Days \\
Parallel Strategy & FSDP (\texttt{HYBRID\_SHARD}) \\
\bottomrule
\end{tabular}
\end{table}

\section{Attention Masking Strategy}
\label{appendix:atten_mask}
\begin{figure}[htbp]
    \centering
    \includegraphics[width=0.7\textwidth,keepaspectratio]{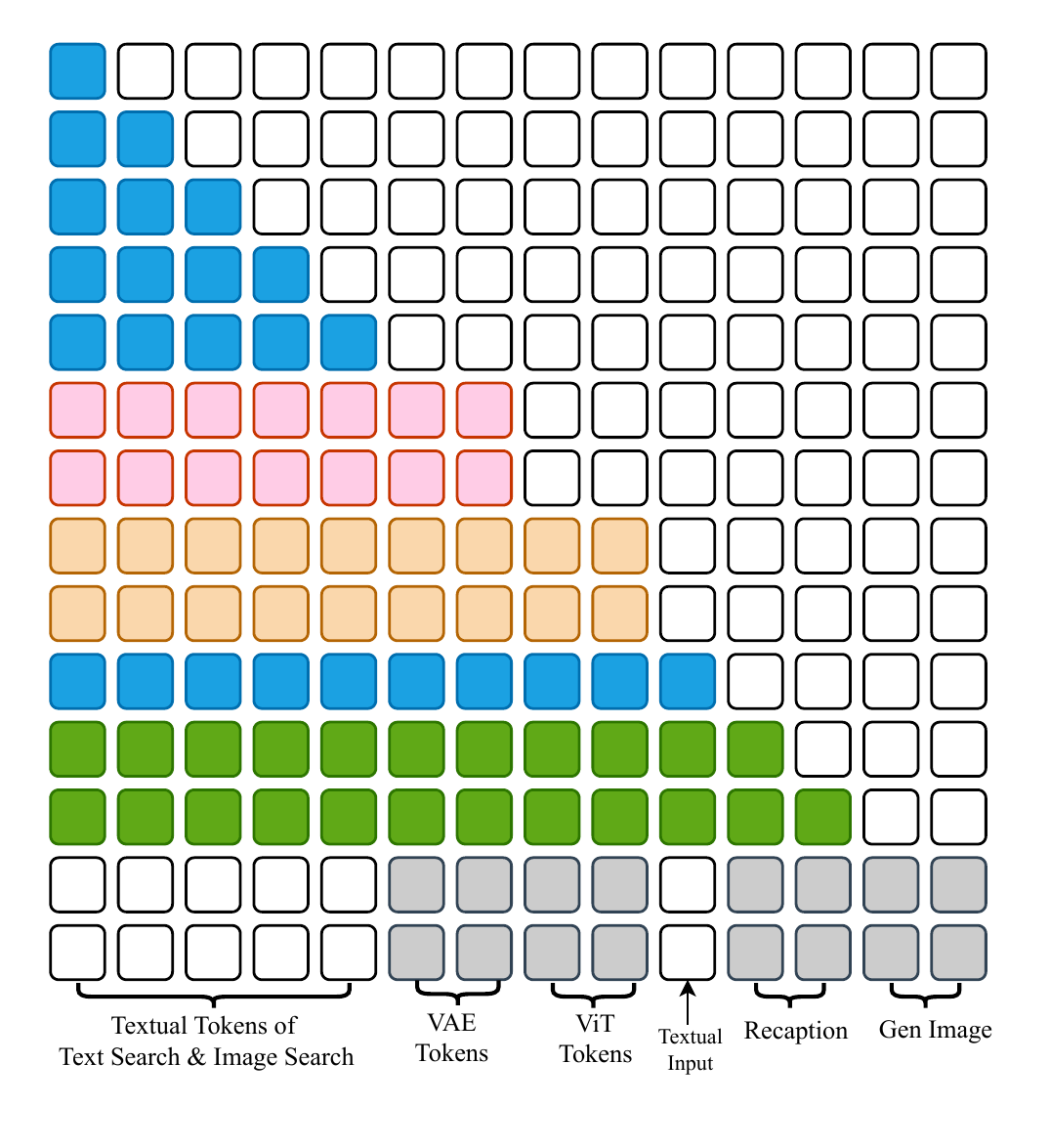}
    \vspace{-20pt}
    \caption{\textbf{Hybrid attention masking strategy for agentic image generation. }The matrix shows causal attention for reasoning traces and textual dialog (blue) and full attention for reference visual-textual evidence (pink/orange/green). Notably, the generation phase (grey) employs a restricted mask that filters out historical reasoning noise, forcing the latent flow-matching tokens to attend only to relevant reference images and recaptions for high-fidelity synthesis.}
    \label{fig:attn}
\end{figure}

We implement a hybrid attention masking strategy tailored for interleaved agentic data to balance sequential logic with precise visual grounding. For the textual components of the sequence, including agentic dialogs and reasoning steps, a standard causal masking is applied to preserve the temporal and logical flow of the conversation. In contrast, retrieved reference images—represented by a composite of VAE and ViT tokens—employ a full attention mechanism, allowing these tokens to interact globally to facilitate holistic visual feature extraction. Notably, during the subsequent image-generation phase, the latent flow-matching tokens (i.e., VAE noisy tokens) are specifically restricted to attending exclusively to the preceding recaption tokens and the retrieved reference image tokens. By selectively masking out irrelevant textual reasoning traces and historical dialogs at this stage, we prevent large segments of high-noise text from interfering with the synthesis process, thereby ensuring that the generated visual content remains precisely grounded in the condensed recaptioning and the provided visual evidence. See Fig. ~\ref{fig:attn} for details of our attention masking strategy.

\section{Evaluation Protocols}
\label{appen:eval_protocols}

To ensure rigorous and standardized evaluation across benchmarks with distinct task formulations and assessment objectives, we adhere to the official evaluation protocols of each benchmark whenever possible. For all existing benchmarks, we preserve their original MLLM-based evaluation settings, including the designated expert evaluators and scoring schemes, to maintain consistency with prior literature and improve the fairness and comparability of our results.

\paragraph{WiSE Evaluation.}
For the \textbf{WiSE Benchmark}, we employ \textbf{GPT-4o} \citep{openai2024gpt4ocard} as the expert evaluator to assess the alignment between generated images and world knowledge. Given the textual prompt and the model-generated result, the evaluator is instructed to score the generation from three complementary aspects: \textit{Consistency}, \textit{Realism}, and \textit{Aesthetic Quality}. Specifically, \textit{Consistency} evaluates the accuracy and completeness with which the generated image reflects the prompt; \textit{Realism} assesses the physical plausibility, including adherence to physical laws and accurate material representation; and \textit{Aesthetic Quality} measures the overall artistic appeal, including composition and color harmony. Each dimension is evaluated on a discrete scale from 0 to 2 (0 for Rejected, 1 for Conditional, and 2 for Exemplary).

To obtain a unified measure of knowledge-image alignment, we define the overall metric, \textbf{WiScore}, as a weighted combination of the three dimensions:
\begin{equation}
\text{WiScore} =
\alpha_1 \cdot \textbf{Consistency}
+ \alpha_2 \cdot \textbf{Realism}
+ \alpha_3 \cdot \textbf{Aesthetic Quality},
\end{equation}
\begin{equation}
\text{subject to} \quad \alpha_1 + \alpha_2 + \alpha_3 = 1.
\end{equation}
In our experiments, we set $\alpha_1 = 0.7$, $\alpha_2 = 0.2$, and $\alpha_3 = 0.1$. This configuration explicitly prioritizes \textit{Consistency}, reflecting that accurately representing the intended objects and relationships grounded in world knowledge is the most critical criterion. Meanwhile, \textit{Realism} and \textit{Aesthetic Quality} are retained as auxiliary dimensions to ensure overall visual excellence without overshadowing the core knowledge assessment.

\paragraph{KiTTEN Evaluation.}
For the \textbf{KiTTEN Benchmark}, we utilize \textbf{GPT-4o} \citep{openai2024gpt4ocard} as the expert evaluator to systematically assess the fine-grained visual fidelity of generated entities. Given the textual prompt, ground-truth reference images of the target entity, and the model-generated result, the evaluation framework decouples the assessment into two primary dimensions: \textit{Text Alignment} and \textit{Entity Alignment}. Specifically, \textit{Text Alignment} evaluates the degree to which the generated image faithfully adheres to the prompt instructions beyond just the entity; and \textit{Entity Alignment} assesses how well the generated image captures precise visual details and realistically resembles the target entity based on the provided reference images. Each dimension is independently scored on an integer scale from 1 to 5, where 1 indicates no alignment and 5 denotes complete faithfulness. This decoupled design deliberately avoids fusing the scores into a single metric, allowing for a nuanced analysis of the trade-offs between preserving specific entity knowledge and strictly following complex textual constraints.

\paragraph{T2I-FactBench Evaluation.} 
For the T2I-FactBench, we use a multi-round VQA protocol with GPT-4o across three levels: SKCM, SKCI, and MKCC. The evaluation consists of three progressive stages. 

In the first round, \textbf{Concept Factuality} assesses the generated knowledge concept $c_i$ in image $I_i$ against its reference $I^R_i$ across four dimensions: Shape ($S$), Color ($C$), Texture ($T$), and Feature Details ($F$). For each dimension, GPT-4o assigns a binary score (0 or 1). To account for multiple concepts at the MKCC level, the score is defined as:
\begin{equation}
    \text{Concept Factuality} = \frac{1}{N_i} \sum_{j=1}^{N_i} \left( \frac{S_{ij} + C_{ij} + T_{ij} + F_{ij}}{4} \right),
    \label{eq: concept_factuality}
\end{equation}
where $N_i$ denotes the number of concepts within $I_i$, and $S_{ij}, C_{ij}, T_{ij}, F_{ij}$ represent binary scores for the $j$-th concept. 

The second round, \textbf{Instantiation Completeness}, verifies whether concept $c$ is present and if the instantiation phrase $p$ is successfully realized. A score of 1 is assigned only if both conditions are satisfied. 

Finally, for the MKCC level, a third round evaluates \textbf{Composition Factuality} across four dimensions: Seamless Transition ($S_i$), Visual Completeness ($V_i$), Authenticity ($A_i$), and Prompt Following ($P_i$). The composition factuality score is defined as:
\begin{equation}
    \text{Composition Factuality} = \frac{S_{i} + V_{i} + A_{i} + P_{i}}{4},
    \label{eq: Composition}
\end{equation}
where $S_{i}, V_{i}, A_{i}, P_{i} \in \{0, 1\}$ represent scores for the respective dimensions for the $i$-th prompt.

\paragraph{FactIP Benchmark Evaluation.}

\label{appendix: factip_eval}

For our proposed \textbf{FactIP Benchmark}, we employ \textbf{Seed2.0} \citep{seedseed2} as the expert evaluator to conduct a fine-grained and structured assessment of generated images and the evaluation prompt can be found in \ref{subsec:Evluation_prompt}. Given the textual prompt, two reference images (\textbf{GT1} and \textbf{GT2}), and the model-generated result (\textbf{AS}), the evaluator is instructed to jointly compare the generated image against both the prompt and the references, and to score it from four complementary aspects: \textit{Clarity}, \textit{Content}, \textit{Aesthetics}, and \textit{Relevance}. Specifically, \textit{Clarity} evaluates image sharpness, visual cleanliness, and the richness of perceptual details; \textit{Content} measures whether the generated image faithfully captures the key semantic elements and compositional requirements specified in the prompt; \textit{Aesthetics} assesses the overall visual quality, including composition, lighting, color harmony, and stylistic appeal; and \textit{Relevance} measures whether the generated image preserves the intended IP identity and remains consistent with the defining attributes inferred from the two reference images. Each dimension is scored on an integer scale from 0 to 10 by the expert evaluator. Following the evaluation protocol, the two references (\textbf{GT1} and \textbf{GT2}) are considered jointly, such that stable identity traits are emphasized while variations already existing across the references are not over-penalized.

To obtain a unified measure of generation quality, we define the overall score as a weighted combination of the four dimensions:
\begin{equation}
\text{Overall Score} =
\alpha_1 \cdot \textbf{Clarity}
+ \alpha_2 \cdot \textbf{Content}
+ \alpha_3 \cdot \textbf{Aesthetics}
+ \alpha_4 \cdot \textbf{Relevance},
\end{equation}
\begin{equation}
\text{subject to} \quad \alpha_1 + \alpha_2 + \alpha_3 + \alpha_4 = 1.
\end{equation}
In our experiments, we set $\alpha_1 = 0.05$, $\alpha_2 = 0.10$, $\alpha_3 = 0.10$, and $\alpha_4 = 0.75$. This configuration explicitly prioritizes \textit{Relevance}, reflecting that consistency with the actual IP identity is the most critical criterion in FactIP Benchmark. Such a design is also aligned with our emphasis on broad world knowledge utilization, since accurately preserving the target IP requires the model to recognize and faithfully express knowledge-intensive identity cues rather than merely producing visually plausible content. Meanwhile, \textit{Clarity}, \textit{Content}, and \textit{Aesthetics} are retained as important auxiliary dimensions to ensure that the final metric also captures perceptual quality and semantic completeness. Finally, the weighted overall score is multiplied by 10, so that the final score range is normalized to \([0, 100]\), where a higher score indicates better IP fidelity and overall generation quality. More Evaluations showcases can be found in \ref{app: factip_showcases}.

\section{More Details about FactIP benchmark}

\begin{figure}[htbp]
    \centering
    \includegraphics[width=1.00\textwidth,keepaspectratio]{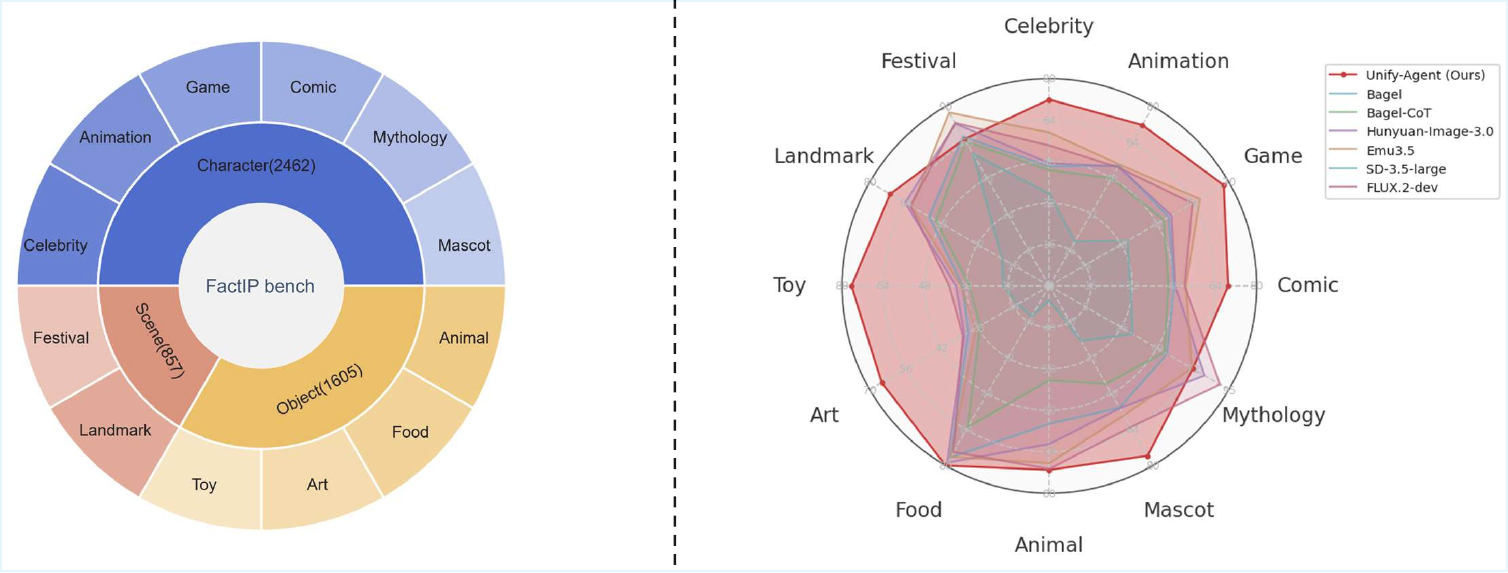}
    \vspace{-20pt}
    \caption{(a) Hierarchical category distribution of FactIP Bench, consisting of three major groups (Character, Scene, and Object) and 12 fine-grained subcategories. (b) Category-wise comparison of different methods on FactIP Bench, where the radar chart presents the overall scores across all subcategories.}
    \label{fig:factip_construction}
    \vspace{-20pt}
\end{figure}

\begin{table*}[!t]
\centering
\footnotesize
\caption{Detailed category statistics of our benchmark. The full benchmark contains three major categories and 12 fine-grained subcategories, totaling 2{,}462 prompts.}
\label{tab:highlevel_category_statistics}
\setlength{\tabcolsep}{8pt}
\renewcommand{\arraystretch}{1.3}
\resizebox{\linewidth}{!}{
\begin{tabular}{c l p{11cm} r}
\toprule
\textbf{Category} & \textbf{Subcategory} & \textbf{Description} & \textbf{Num} \\
\midrule
\multirow{6}{*}{\textbf{CHARACTER}}
& Animation & Animated characters, creatures, equipment, and iconic locations from anime and animated media. & 438 \\
\cmidrule(lr){2-4}
& Comic & Characters and visual elements originating from comic books and manga series. & 363 \\
\cmidrule(lr){2-4}
& Celebrity & Prominent figures across diverse domains, including scientists, political leaders, business executives, athletes, and entertainment personalities. & 300 \\
\cmidrule(lr){2-4}
& Game & Video game characters, weapons, equipment, and other in-game visual elements. & 272 \\
\cmidrule(lr){2-4}
& Mascot & Official mascots representing Olympic Games, regional events, and corporate brands. & 77 \\
\cmidrule(lr){2-4}
& Mythology & Universally recognized mythological narratives and legendary figures, e.g., Kuafu Chasing the Sun. & 50 \\
\midrule
\multirow{4}{*}{\textbf{OBJECT}}
& Food & Cuisines, regional delicacies, desserts, and beverages with cultural significance. & 316 \\
\cmidrule(lr){2-4}
& Cultural Relic / Art & National treasures, classical calligraphy, paintings, sculptures, and fine art pieces. & 126 \\
\cmidrule(lr){2-4}
& Toy & Collectible figures, designer toys, and model kits with cultural relevance, e.g., Labubu. & 123 \\
\cmidrule(lr){2-4}
& Animal / Plant & Individually notable animals and plants with distinct public recognition, e.g., Giant Panda Qizai. & 50 \\
\midrule
\multirow{2}{*}{\textbf{SCENE}}
& Landmark & Renowned scenic spots, architectural landmarks, monuments, and heritage sites. & 297 \\
\cmidrule(lr){2-4}
& Festival / Celebration & Visual elements and symbols associated with well-known festivals and cultural celebrations. & 50 \\
\bottomrule
\end{tabular}}
\end{table*}

\begin{table*}[t]
    \caption{
    Per-subtask overall scores on the FactIP benchmark.
    \textbf{Character} includes celebrity, animation, game, comic, mythology, and mascot prompts;
    \textbf{Object} includes animals/plants, food, cultural relic/art, and toy prompts;
    \textbf{Scene} includes landmark and celebration prompts.
    \colorbox{purple!30}{\phantom{xx}} denotes the best score and \colorbox{purple!15}{\phantom{xx}} the second-best within each group.
    }
    \centering
    \footnotesize
    \setlength{\tabcolsep}{4pt}
    \resizebox{\textwidth}{!}{
    \begin{tabular}{c cccccc cccc cc c}
    \toprule
    \multirow{2}{*}{\centering \textbf{Model}}
    & \multicolumn{6}{c}{\textbf{Character}}
    & \multicolumn{4}{c}{\textbf{Object}}
    & \multicolumn{2}{c}{\textbf{Scene}}
    & \multirow{2}{*}{\centering \textbf{Overall}} \\
    \cmidrule(lr){2-7} \cmidrule(lr){8-11} \cmidrule(lr){12-13}
    & Celebrity & Animation & Game & Comic & Mythology & Mascot & Animal & Food & Art & Toy & Landmark & Festival & \\
    \midrule
    \rowcolor{blue!5}
    \multicolumn{14}{c}{\textbf{Commercial Models}} \\
    GPT-Image-1 & 77.0 & 63.1 & 66.4 & 60.3 & \cellcolor{purple!15}{95.7} & 64.7 & \cellcolor{purple!30}\textbf{89.0} & 88.6 & 40.9 & 52.2 & 76.1 & 94.2 & 69.3 \\
    GPT-Image-1.5 & 77.7 & 64.0 & 64.1 & 61.9 & 93.0 & 69.4 & \cellcolor{purple!15}{88.1} & 92.1 & 42.8 & 48.9 & 76.0 & \cellcolor{purple!15}{95.4} & 69.9 \\
    Seedream-4.5 & 86.0 & 75.6 & 82.3 & 80.8 & 88.9 & 88.1 & 77.0 & 92.4 & 64.8 & 83.0 & 83.5 & 88.9 & 82.0 \\
    Seedream-4 & 86.8 & 76.1 & 85.3 & 81.0 & 88.9 & 85.2 & 80.1 & 91.4 & 70.2 & 83.8 & 85.8 & 87.5 & 83.0 \\
    Seedream-5 & \cellcolor{purple!15}{87.5} & \cellcolor{purple!30}\textbf{85.0} & \cellcolor{purple!30}\textbf{89.5} & \cellcolor{purple!30}\textbf{86.8} & 87.8 & \cellcolor{purple!30}\textbf{90.9} & 81.5 & \cellcolor{purple!30}\textbf{96.0} & 77.5 & \cellcolor{purple!15}{89.5} & \cellcolor{purple!30}\textbf{88.8} & 82.8 & \cellcolor{purple!15}{87.3} \\
    Qwen-Image-2.0-Pro & 68.8 & 64.5 & 71.9 & 64.6 & 80.5 & 74.3 & 75.5 & 82.7 & 65.0 & 79.3 & 80.2 & 84.0 & 71.1 \\
    Nano Banana & 48.8 & 53.2 & 57.8 & 49.0 & 70.0 & 65.5 & 49.4 & 64.2 & \cellcolor{purple!15}{85.2} & 69.1 & 59.4 & 56.7 & 56.6 \\
    Nano Banana-Pro & 75.0 & 61.4 & 62.9 & 54.6 & \cellcolor{purple!30}\textbf{97.0} & 63.6 & 87.0 & 89.9 & 43.8 & 46.3 & 72.3 & 90.3 & 66.7 \\
    Nano Banana-2 & \cellcolor{purple!30}\textbf{92.5} & \cellcolor{purple!15}{82.6} & \cellcolor{purple!15}{89.0} & \cellcolor{purple!15}{83.7} & 95.2 & \cellcolor{purple!15}{89.1} & \cellcolor{purple!15}{88.1} & \cellcolor{purple!15}{94.8} & \cellcolor{purple!30}\textbf{86.9} & \cellcolor{purple!30}\textbf{94.8} & \cellcolor{purple!15}{88.3} & \cellcolor{purple!30}\textbf{96.4} & \cellcolor{purple!30}\textbf{88.5} \\
    \rowcolor{blue!5}
    \multicolumn{14}{c}{\textbf{Generation Only}} \\
    Pixel-Art-XL & 24.7 & 5.6 & 12.1 & 12.1 & 19.8 & 5.8 & 25.9 & 1.4 & 2.2 & 5.0 & 6.8 & 26.1 & 12.1 \\
    FLUX.1-schnell & 26.0 & 16.8 & 25.9 & 30.1 & 50.0 & 17.4 & 49.6 & 11.8 & 12.1 & 18.9 & 19.7 & 54.9 & 24.0 \\
    SD-3-medium & 33.7 & 17.8 & 23.8 & 31.5 & 49.1 & 18.7 & 49.0 & 13.1 & 11.3 & 16.6 & 19.2 & 58.9 & 25.7 \\
    SDXL-base-0.9 & 38.2 & 18.7 & 27.5 & 29.9 & 40.8 & 15.3 & 43.2 & 11.5 & 11.8 & 17.4 & 18.7 & 60.1 & 26.5 \\
    SD-3.5-medium & 33.4 & 19.7 & 27.2 & 32.9 & 50.7 & 16.2 & 47.0 & 8.8 & 11.1 & 16.8 & 21.2 & 64.3 & 26.5 \\
    SD-3.5-large & 35.6 & 19.8 & 30.4 & 31.3 & 45.7 & 24.6 & 42.8 & 13.7 & 12.3 & 17.8 & 21.0 & 66.7 & 27.5 \\
    FLUX.1-dev & 36.1 & 21.4 & 28.1 & 31.9 & 57.7 & 22.6 & 50.9 & 18.0 & 15.8 & 20.9 & 23.2 & 58.8 & 28.9 \\
    Z-Image & 49.6 & 52.7 & \cellcolor{purple!15}{54.7} & \cellcolor{purple!15}{51.3} & 67.6 & 58.1 & \cellcolor{purple!30}\textbf{77.3} & \cellcolor{purple!30}\textbf{75.4} & 32.1 & 34.9 & \cellcolor{purple!15}{62.4} & \cellcolor{purple!15}{78.2} & 54.2 \\
    FLUX.2-dev & \cellcolor{purple!30}\textbf{54.2} & \cellcolor{purple!15}{52.9} & \cellcolor{purple!30}\textbf{55.8} & \cellcolor{purple!30}\textbf{52.2} & \cellcolor{purple!30}\textbf{90.4} & \cellcolor{purple!30}\textbf{62.9} & \cellcolor{purple!15}{75.3} & \cellcolor{purple!15}{74.0} & \cellcolor{purple!15}{33.7} & \cellcolor{purple!30}\textbf{38.6} & 61.9 & \cellcolor{purple!30}\textbf{81.9} & \cellcolor{purple!30}\textbf{56.3} \\
    Qwen-Image & \cellcolor{purple!15}{52.5} & \cellcolor{purple!30}\textbf{56.3} & 52.2 & 50.8 & \cellcolor{purple!15}{79.0} & \cellcolor{purple!15}{62.5} & 68.8 & 69.2 & \cellcolor{purple!30}\textbf{33.9} & \cellcolor{purple!15}{37.9} & \cellcolor{purple!30}\textbf{65.8} & 73.3 & \cellcolor{purple!15}{55.4} \\
    \midrule
    \rowcolor{blue!5}
    \multicolumn{14}{c}{\textbf{Unified MLLM}} \\
    Janus-1.3B & 16.4 & 18.4 & 23.5 & 26.2 & 25.7 & 13.3 & 38.6 & 13.8 & 3.8 & 9.6 & 20.2 & 33.7 & 19.3 \\
    Janus-Pro & 23.0 & 31.6 & 33.1 & 35.5 & 61.1 & 27.3 & 43.5 & 39.1 & 9.3 & 16.7 & 30.3 & 45.7 & 30.3 \\
    Emu3 & 30.6 & 28.6 & 31.7 & 33.8 & 36.5 & 21.3 & 45.6 & 26.1 & 15.5 & 21.3 & 30.9 & 49.0 & 30.0 \\
    Emu3.5 & \cellcolor{purple!15}{59.2} & \cellcolor{purple!15}{54.1} & \cellcolor{purple!15}{58.7} & \cellcolor{purple!15}{52.7} & 76.0 & \cellcolor{purple!15}{57.7} & \cellcolor{purple!15}{74.2} & 76.2 & 28.7 & 34.1 & \cellcolor{purple!15}{61.3} & \cellcolor{purple!30}\textbf{87.0} & \cellcolor{purple!15}{57.2} \\
    Hunyuan-Image-3.0 & 47.4 & 53.2 & 47.6 & 48.6 & \cellcolor{purple!30}\textbf{82.5} & 54.5 & 70.5 & \cellcolor{purple!15}{78.8} & \cellcolor{purple!15}{33.1} & \cellcolor{purple!15}{36.0} & \cellcolor{purple!15}{64.2} & \cellcolor{purple!15}{81.5} & 53.4 \\
    Echo4o & 44.9 & 48.1 & 43.3 & 43.5 & 66.5 & 52.1 & 60.5 & 62.2 & 27.7 & 30.3 & 55.1 & 70.0 & 47.3 \\
    Bagel & 46.1 & 53.7 & 46.5 & 48.2 & 63.3 & 54.3 & 66.6 & 76.5 & 31.6 & 33.5 & 53.5 & 75.2 & 50.9 \\
    Bagel-CoT & 44.7 & 48.0 & 44.6 & 46.0 & 62.0 & 43.7 & 58.2 & 63.0 & 27.0 & 30.8 & 50.5 & 72.8 & 47.0 \\
    \textbf{Unify-Agent (Ours)} & \cellcolor{purple!30}\textbf{71.9} & \cellcolor{purple!30}\textbf{71.7} & \cellcolor{purple!30}\textbf{68.0} & \cellcolor{purple!30}\textbf{69.0} & \cellcolor{purple!15}{76.5} & \cellcolor{purple!30}\textbf{75.7} & \cellcolor{purple!30}\textbf{75.5} & \cellcolor{purple!30}\textbf{80.2} & \cellcolor{purple!30}\textbf{65.3} & \cellcolor{purple!30}\textbf{76.3} & \cellcolor{purple!30}\textbf{70.9} & 74.0 & \cellcolor{purple!30}\textbf{71.7} \\
    \bottomrule
    \end{tabular}
    }
    \label{tab:factip-appendix}
    \end{table*}

\subsection{Construction}
\label{appendix: factip_construction}
To rigorously evaluate the capability of generative models in synthesizing high-fidelity intellectual property (IP) and culturally significant concepts, we construct the FactIP benchmark. As illustrated in Figure~\ref{fig:factip_construction}(a), the benchmark is meticulously organized into a hierarchical taxonomy. It encompasses three overarching domains—Character, Object, and Scene—which are further decomposed into fine-grained subcategories. This comprehensive structure ensures a diverse and representative evaluation of a model's factual knowledge grounding. For a detailed statistical breakdown and specific descriptions of each subcategory, please refer to Table~\ref{tab:highlevel_category_statistics}. In total, the dataset comprises 2,462 meticulously curated prompts, challenging models to generate recognizable entities ranging from global celebrities and intricate art toys to iconic landmarks. To facilitate efficient and accessible model evaluation, we additionally construct a lightweight version, FactIP-Mini, by sampling 500 prompts from the full dataset while strictly preserving the original category distribution.

\subsection{Results}

We present a comprehensive performance analysis of various leading generative models on the FactIP benchmark, with all reported results evaluated on the representative FactIP-Mini dataset. Figure~\ref{fig:factip_construction}(b) provides a visual category-wise comparison, highlighting the relative strengths and weaknesses of different architectures across the primary evaluation dimensions. The radar chart succinctly captures the overarching performance trends, revealing that Unified MLLMs and state-of-the-art commercial models generally exhibit superior IP fidelity and factual consistency compared to earlier generation-only paradigms. A detailed, quantitative breakdown of the per-subtask overall scores is provided in Table~\ref{tab:factip-appendix}. These fine-grained results emphasize the necessity of advanced multimodal reasoning and extensive factual grounding to accurately render domain-specific visual characteristics. Notably, our proposed Unify-Agent demonstrates highly competitive and balanced performance across all taxonomic branches, underscoring the efficacy of its unified architectural design in handling complex, knowledge-intensive visual generation tasks.

\section{Prompt}
\label{sec:prompt}

\subsection{System Prompt}
\label{subsec:system_prompt}

\begin{tcolorbox}[
    colback=lightgray!10,
    colframe=black,
    title={\textbf{SYSTEM\_PROMPT}},
    breakable
]
\vspace{0.5em}
\begin{Verbatim}[breaklines=true, breaksymbol={}, fontsize=\small]
    You are helping to build a high-quality visual generation dataset.
    Your task is to gather information and reference images for creating detailed image descriptions.

    Your Goal: Create a detailed recaption for "{image_prompt}"
    about the IP "{ip_name}" ({country}).

    Natural Workflow (think step by step):
    1. First, search for background information about this IP/character to understand who they are, their characteristics, style, and context. This knowledge will help you craft more accurate image search queries.
    2. Then, search for reference images of this IP/character. Good reference visuals are essential for the final detailed description.
    3. Finally, I will provide you with the downloaded reference images, and you will generate a detailed <recaption> that references "image_1" and "image_2" specifically.

    Tool Call Format (IMPORTANT - use ONLY this format):
    <tool_call>
    {{"name": "tool_name", "arguments": {{"param1": "value1"}}}}
    </tool_call>
    Examples:
    - Text search: <tool_call>{{"name": "text_search", "arguments":
      {{"q": "search query", "hl": "zh", "top_k": 5}}}}</tool_call>
    - Image search: <tool_call>{{"name": "search_image", "arguments":
      {{"q": "image query", "hl": "zh", "num": 8}}}}</tool_call>
\end{Verbatim}
\end{tcolorbox}

\begin{tcolorbox}[
    colback=lightgray!10,
    colframe=black,
    title={\textbf{TOOLS\_DEFINITION}},
    breakable
]
\vspace{0.5em}
\begin{Verbatim}[breaklines=true, breaksymbol={}, fontsize=\small]
    tools = [
        {
            "type": "function",
            "function": {
                "name": "text_search",
                "description": "Search the web for text information
                about a query. Use this to get background information
                about IPs, people, places, or topics.",
                "parameters": {
                    "type": "object",
                    "properties": {
                        "q": {
                            "type": "string",
                            "description": "Search query"
                        },
                        "hl": {
                            "type": "string",
                            "description": "Language code
                            (e.g., 'en', 'zh')",
                            "default": "en"
                        },
                        "top_k": {
                            "type": "integer",
                            "description": "Number of results
                            to return",
                            "default": 5
                        }
                    },
                    "required": ["q"]
                }
            }
        },
        {
            "type": "function",
            "function": {
                "name": "search_image",
                "description": "Search for images on the web. Use
                this to find relevant images about the IP or topic.",
                "parameters": {
                    "type": "object",
                    "properties": {
                        "q": {
                            "type": "string",
                            "description": "Search query for images"
                        },
                        "location": {
                            "type": "string",
                            "description": "Location for search",
                            "default": "United States"
                        },
                        "hl": {
                            "type": "string",
                            "description": "Language code",
                            "default": "en"
                        },
                        "num": {
                            "type": "integer",
                            "description": "Number of images
                            to return",
                            "default": 8
                        }
                    },
                    "required": ["q"]
                }
            }
        }
    ]
\end{Verbatim}
\end{tcolorbox}

\subsection{Evluation Prompt}
\label{subsec:Evluation_prompt}

\begin{tcolorbox}[
    colback=lightgray!10,
    colframe=black,
    title={\textbf{EVALUATION\_PROMPT}},
    breakable
]
\vspace{0.5em}
\begin{Verbatim}[breaklines=true, breaksymbol={}, fontsize=\small]
    You are an image evaluation assistant. Compare AS (assistant-generated image) against GT1 and GT2 (ground-truth images) given the Prompt.

    Your task is to return exactly 6 fields: 5 integer scores (0-10) and 1 rationale string.

    Evaluate these 5 dimensions independently:

    1. "clarity": image sharpness, absence of blur/artifacts/noise, and richness of visible details.
    2. "content_quality": faithfulness to the Prompt, subject completeness, and semantic coherence.
    3. "aesthetics": visual appeal, composition, lighting/color harmony, and style consistency.
    4. "text_relevance_ip": IP identity consistency. This means whether AS preserves the same character/object/IP identity as GT1/GT2 based on distinctive traits (e.g. face, hairstyle, costume, colors, species/object-defining features). Do NOT require exact matching of pose, background, camera angle, or composition.

    Important instructions:
    - Use GT1 and GT2 jointly to infer the stable identity and attributes of the IP.
    - Do not penalize AS for differences that also vary between GT1 and GT2.
    - Score each dimension independently before assigning the overall score.
    - All five score fields must be integers from 0 to 10.
    - "rationale" must be a single short string with at least two concrete evidence points, separated by semicolons.

    You MUST respond with ONLY this exact JSON structure, with all 6 keys present and no extra keys:

    {"clarity": 7, "content_quality": 8, "aesthetics": 7,
     "text_relevance_ip": 8,
     "rationale": "Evidence 1; Evidence 2"}

    Do not use markdown fences.
    Do not output any text before or after the JSON.
    If uncertain, still output the best-effort JSON with all required keys.
\end{Verbatim}
\end{tcolorbox}

\subsection{Judge Prompt}
\label{subsec:judge_prompt}

\begin{tcolorbox}[
    colback=lightgray!10,
    colframe=black,
    title={\textbf{JUDGE\_PROMPT}},
    breakable
]
\vspace{0.5em}
\begin{Verbatim}[breaklines=true, breaksymbol={}, fontsize=\small]
    You are an expert Image Quality Assessor for an AI training
    pipeline. Your task is to evaluate whether a downloaded image
    is a high-quality visual reference for a specific Intellectual
    Property (IP).

    You must rate the image on a scale of 0 to 10 based on the
    following strict rubrics:

    ### 1. IP Consistency (Critical)
    - Pass: The image clearly depicts the specific character/object requested.
    - Fail (Score 0): The image shows a completely different character, a landscape, a real person (cosplay) if the IP is anime, or an unrelated object.

    ### 2. Layout & Composition
    - Subject-Centric: The IP subject must be the main focus.
    - Face Visibility: For characters, the face must be clearly visible (preferably front-facing or 3/4 view). REJECT if the face is tiny, distant, or back-facing.
    - No Text-Heavy: REJECT images that are primarily movie posters with large text overlays, book covers, or infographics where the subject is obscured by text.
    - No Collages: REJECT split-screens, manga panels, or multiple images stitched together. Single-scene images only.

    ### 3. Visual Quality
    - Clarity: The image must be sharp. REJECT if it is
      severely blurry, pixelated, or has heavy jpeg compression
      artifacts.

    ### 4. Watermarks & Obstructions
    - Reject: Large, obstructive watermarks covering the face or main body (e.g., full-screen stock photo watermarks).
    - Accept: Small, unobtrusive logos in the corner are acceptable but lower the score slightly.

    ### Scoring Guide:
    - 0: Wrong IP or completely unusable.
    - 1-5 (Reject): Correct IP but violates a major rule (Blurry, Huge Watermark, Text-Heavy, Collage, Tiny Face).
    - 6-7 (Borderline): Usable but not ideal (Side profile, small corner watermark, medium resolution).
    - 8-10 (Excellent): Perfect reference (High-res, clear front-facing, no text, clean).
\end{Verbatim}
\end{tcolorbox}

\begin{tcolorbox}[
    colback=lightgray!10,
    colframe=black,
    title={\textbf{JUDGE\_Question}},
    breakable
]
\vspace{0.5em}
\begin{Verbatim}[breaklines=true, breaksymbol={}, fontsize=\small]
    Please evaluate the provided image for the IP: "{ip_name}".

    Assess the image based on the system rubrics.
    Return your response in JSON format ONLY with the following
    structure:
    {{
        "score": <int, 0-10>,
        "reason": "<string, a concise explanation of the score,
        mentioning any specific flaws like 'text-heavy',
        'watermark', or 'blurry'>",
        "is_text_heavy": <bool>,
        "has_watermark": <bool>
    }}
\end{Verbatim}
\end{tcolorbox}

\subsection{Prompt-Generation Prompt}
\label{subsec:prompt-generation_prompt}

\begin{tcolorbox}[
    colback=lightgray!10,
    colframe=black,
    title={\textbf{PROMPT-GENERATION\_PROMPT}},
    breakable
]
\vspace{0.5em}
\begin{Verbatim}[breaklines=true, breaksymbol={}, fontsize=\small]
    You are a specialized Image Generation Prompt Expert. Your
    mission is to generate a single, high-quality image generation
    prompt based on a specific IP (person/character). You must
    describe "Who is doing What in Which scene."

## 1. Core Logic & Language Rules (CRITICAL)

Step 1: Determine the IP's Nationality
* Case A: If the IP is Chinese (Mainland, Hong Kong, Taiwan,
  Macau):
    * Prompt Language: Must be Chinese.
    * Tag Language: Must be Chinese.
    * Language Code: `zh`
* Case B: If the IP is NOT Chinese (USA, UK, Japan, Korea,
  Europe, etc.):
    * Prompt Language: Must be English.
    * Tag Language: Must be English.
    * Language Code: `en`

## 2. Prompt Construction Requirements

1.  Content Elements:
    * Subject: The IP's name (and brief identity if needed for context).
    * Scene: A specific, realistic location fitting the IP's profession (e.g., Office, Studio, Stadium, Stage, Cafe, Street).
    * Action: A dynamic verb describing what they are doing (e.g., Interviewing, Singing, coding, running, drinking coffee).
2.  Realism Constraint:
    * The scene and action must align with the IP's public persona or profession.
    * Do not hallucinate impossible scenarios unless the IP is a fictional fantasy character.
3.  Syntactic Diversity (Avoid Repetition):
    * Do not always use the structure "Name is doing X in Y".
    * Vary your sentence structures:
        * *Scene-first*: "In the [Scene], [Name] is [Action]..."
        * *Action-focused*: "[Name] is [Action] while located
          in [Scene]..."
        * *Descriptive*: "Surrounded by [Context], [Name] is
          [Action]..."

## 3. Output Format

You must output ONLY the XML tags below. Do not output markdown code blocks (like ```xml), explanations, or conversational filler.

<Image_Prompt>The full descriptive sentence</Image_Prompt>
<Tag_Name>Category of the action (e.g., Interview, Speech,
Daily Life)</Tag_Name>
<Language>zh OR en</Language>

*Note on <Tag_Name>*: If Language is `zh`, the tag must be Chinese. If Language is `en`, the tag must be English (e.g., "Hosting", "Street Snap").

REMEMBER: Output ONLY the three XML tags above, nothing else.

## 4. Examples

Input: Robert Downey Jr. (American Actor)
Output:
<Image_Prompt>Sitting in a relaxed pose on a Hollywood talk
show set, Robert Downey Jr. is laughing while telling a
story</Image_Prompt>
<Tag_Name>Interview</Tag_Name>
<Language>en</Language>

Input: Gordon Ramsay (British Chef)
Output:
<Image_Prompt>Gordon Ramsay is carefully plating a gourmet dish
in a busy high-end restaurant kitchen</Image_Prompt>
<Tag_Name>Cooking</Tag_Name>
<Language>en</Language>
\end{Verbatim}
\end{tcolorbox}

\subsection{Recaption Prompt}
\label{subsec:recaption_prompt}

\begin{tcolorbox}[
    colback=lightgray!10,
    colframe=black,
    title={\textbf{RECAPTION\_PROMPT}},
    breakable
]
\vspace{0.5em}
\begin{Verbatim}[breaklines=true, breaksymbol={}, fontsize=\small]
    You are a professional visual language reasoning assistant.
    Your task is to generate a reasoning process and a final
    detailed image description based on two reference images
    (image_1, image_2), the original instruction, and text
    search results.

Input Information
1. Reference Images: Explicitly labeled Reference Image 1 (refer to as image_1) and Reference Image 2 (refer to as image_2).
2. Original Instruction: The user's short request.
3. Background Info: Text information from previous search steps.

Output Format
Strictly follow XML format:
<think>
[Deep reasoning here:
 1. Analyze visual features of image_1 and image_2.
 2. Combine with background info to plan the fusion.
 3. Explicitly state what comes from image_1 and what comes
    from image_2.]
</think>
<recaption>
[The final detailed image description, including
"Scene Description" and "Preservation Statement"]
</recaption>

Core Rules & Constraints
1. Reference Principle: You MUST strictly use "image_1" and "image_2" to refer to the images. DO NOT use vague terms like "the first image" or "reference picture".

2. Descriptive Style: The content of <Instruction> must be a description of the final result (Descriptive), NOT an editing command (Imperative).
   - BAD: "Please put the man from image_1 on the left..."
   - GOOD: "In this realistic outdoor portrait, the man from image_1 stands on the left..."

3. Preservation Statement: At the end of the description, you MUST explicitly state what specifically is preserved from image_1 and image_2.
   - Must include phrases like: "The final image completely preserves [features] from image_1..." and "The final image fully retains [features] from image_2...".
   - Facial Features Preservation (CRITICAL): If image_1 and/or image_2 contain identifiable persons, you MUST preserve their exact facial features as shown in the reference images. Include detailed descriptions such as: 
   face shape, eyebrow shape, eye characteristics, facial  expression, skin tone/complexion, hairstyle, and any distinctive facial features. Use the format: "Preserve the exact facial features of [person name/description] as shown in image_1 and image_2: [detailed facial feature description]. Maintain [their/his/her] [appearance/clothing/style] as referenced in both images."

4. Language Consistency (MANDATORY): The content of <recaption> MUST be written entirely in the SAME language as the original instruction. If the original instruction is in Chinese, ALL descriptions in <recaption> must be in Chinese - NO English allowed. If the original instruction is in English, use English throughout. NEVER mix languages within the same description.

Start the task. Output <think> and <recaption>.
\end{Verbatim}
\end{tcolorbox}

\subsection{Summary Prompt}
\label{subsec:summary_prompt}

\begin{tcolorbox}[
    colback=lightgray!10,
    colframe=black,
    title={\textbf{SUMMARY\_PROMPT}},
    breakable
]
\vspace{0.5em}
\begin{Verbatim}[breaklines=true, breaksymbol={}, fontsize=\small]
Based on the following webpage content, provide a concise summary that is relevant to the query: "{query}"

Webpage Title: {title}
Content:
{content[:2000]}

Please provide a focused summary (2-3 sentences maximum) that directly addresses the query. Focus on the most relevant information.

You MUST format your response using the following structure:
<think>
[Your thinking process about what information is most relevant to the query]
</think>

<response>
[Your concise summary here - 2-3 sentences maximum]
</response>
\end{Verbatim}
\end{tcolorbox}

\clearpage
\section{Image Generation Showcases}
\label{sec:cases}

\begin{figure}[!t]
    \centering
    \includegraphics[width=1.00\textwidth,height=1.0\textheight,keepaspectratio]{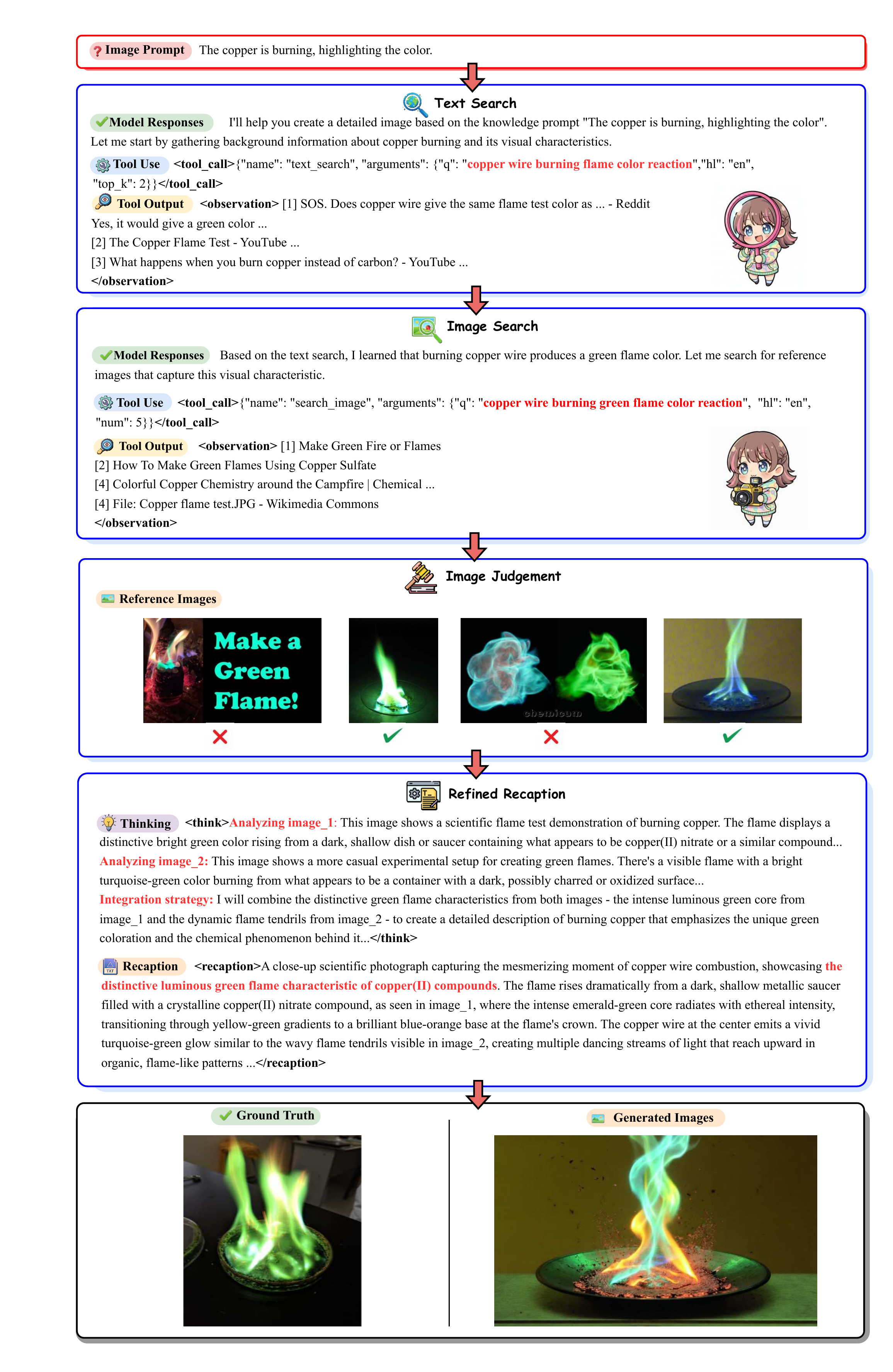}
    \vspace{-20pt}
    \captionsetup{justification=centering}
    \caption{Image generated for the prompt: "The copper is burning, highlighting the color"}
    \label{fig:copper_burning}
    \vspace{-40pt}
\end{figure}

\begin{figure}[!t]
    \centering
    \includegraphics[width=1.00\textwidth,height=1.0\textheight,keepaspectratio]{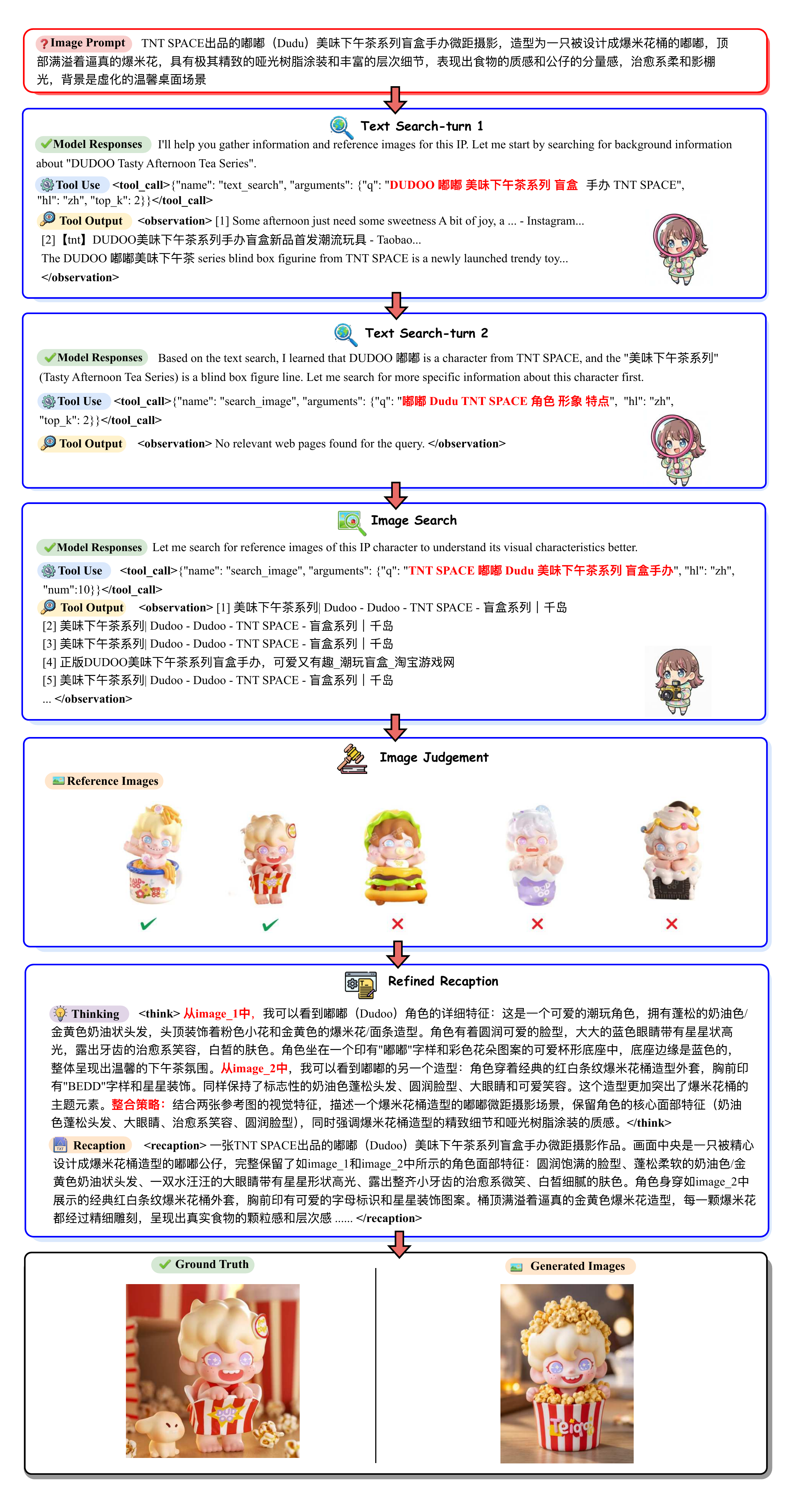}
    \vspace{-20pt}
    \captionsetup{justification=centering}
    \caption{Image generated for the "DUDOO" art toy.}
    \label{fig:art_toy}
    \vspace{-40pt}
\end{figure}

\paragraph{Case 1: Science - Copper Combustion}
This case details the generation of a scientifically accurate image for the knowledge-based prompt \textbf{The copper is burning, highlighting the color}. Recognizing the specific chemical knowledge required, the model first initiated a text search to verify the characteristic flame test reaction of copper, confirming it produces a distinctive green glow. To capture the nuanced visual dynamics, it then performed a targeted image search for "copper wire burning green flame." Through a rigorous image judgment process, the model filtered out low-quality or irrelevant references, selecting high-quality images that showcased both an intense emerald-green core and organic flame tendrils. During the refined recaptioning stage, the model synthesized these visual elements—integrating the luminous intensity of copper(II) compounds with the fluid motion of turquoise-green flames—while adding professional photographic descriptors, providing precise guidance for the generation, accurately depicting the mesmerizing chemical phenomenon. See Fig. ~\ref{fig:copper_burning} for details.
\paragraph{Case 2: Art Toy - DUDOO Tasty Afternoon Tea Series}
This case details the image generation process for the \textbf{DUDOO Tasty Afternoon Tea Series}. The model initially performed a round of text search to establish the general background of the series; however, as the first round yielded only high-level information about the product line, a second round was conducted to specifically probe for the visual identity and iconic character traits of "Dudu." This iterative search strategy compensated for the initial information gap, ensuring a comprehensive understanding of the IP's design elements before sourcing reference images. By integrating these refined character details with visual cues from the retrieved images, the model synthesized an IP-accurate recaption that vividly depicts the "popcorn bucket" theme and matte resin texture. See Fig. ~\ref{fig:art_toy} for details.

\begin{figure}[!t]
    \centering
    \includegraphics[width=1.00\textwidth,height=1.0\textheight,keepaspectratio]{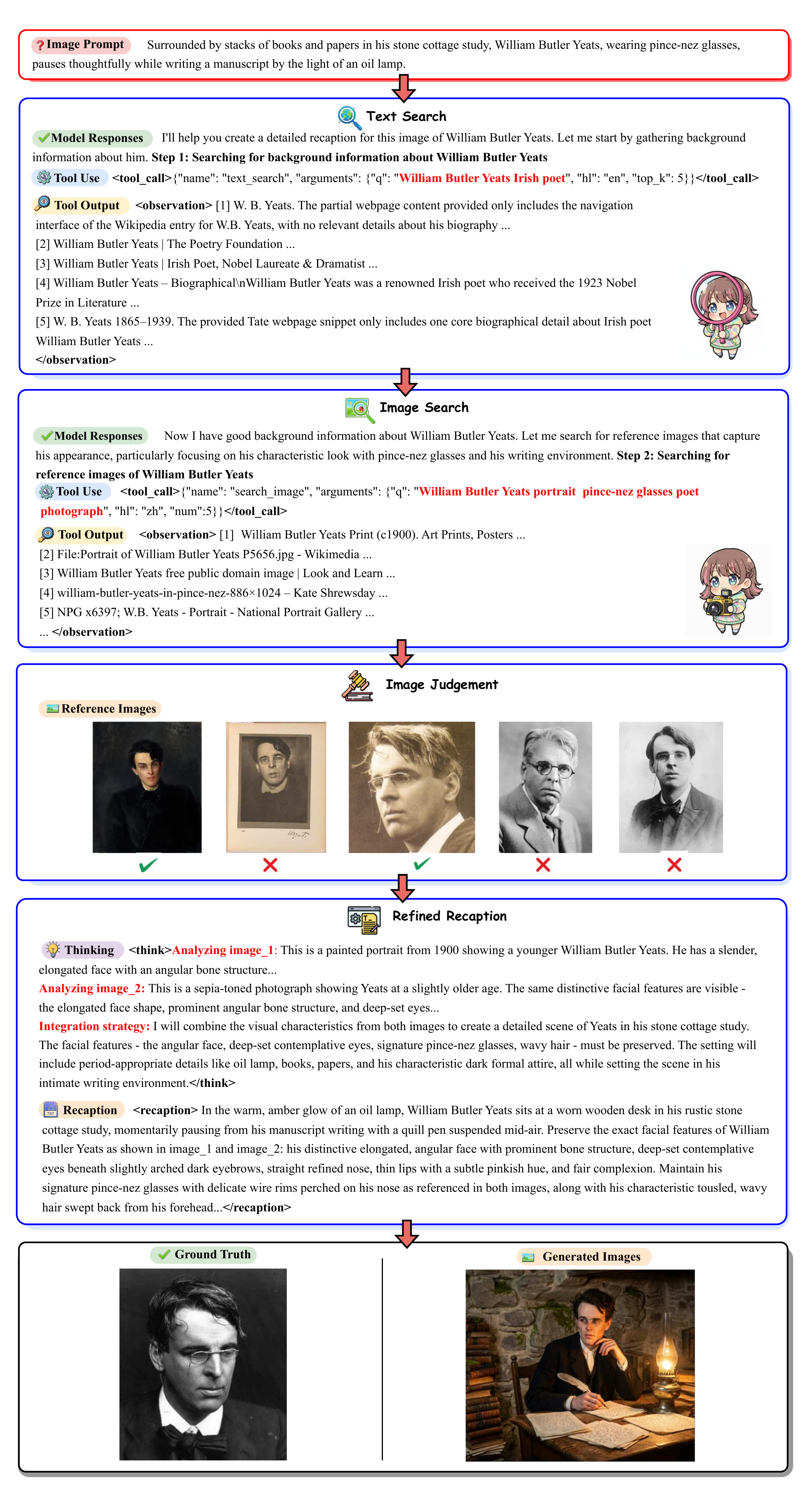}
    \vspace{-20pt}
    \captionsetup{justification=centering}
    \caption{Image generated for William Butler Yeats writing a manuscript.}
    \label{fig:yeets}
    \vspace{-40pt}
\end{figure}

\begin{figure}[!t]
    \centering
    \includegraphics[width=1.00\textwidth,height=1.0\textheight,keepaspectratio]{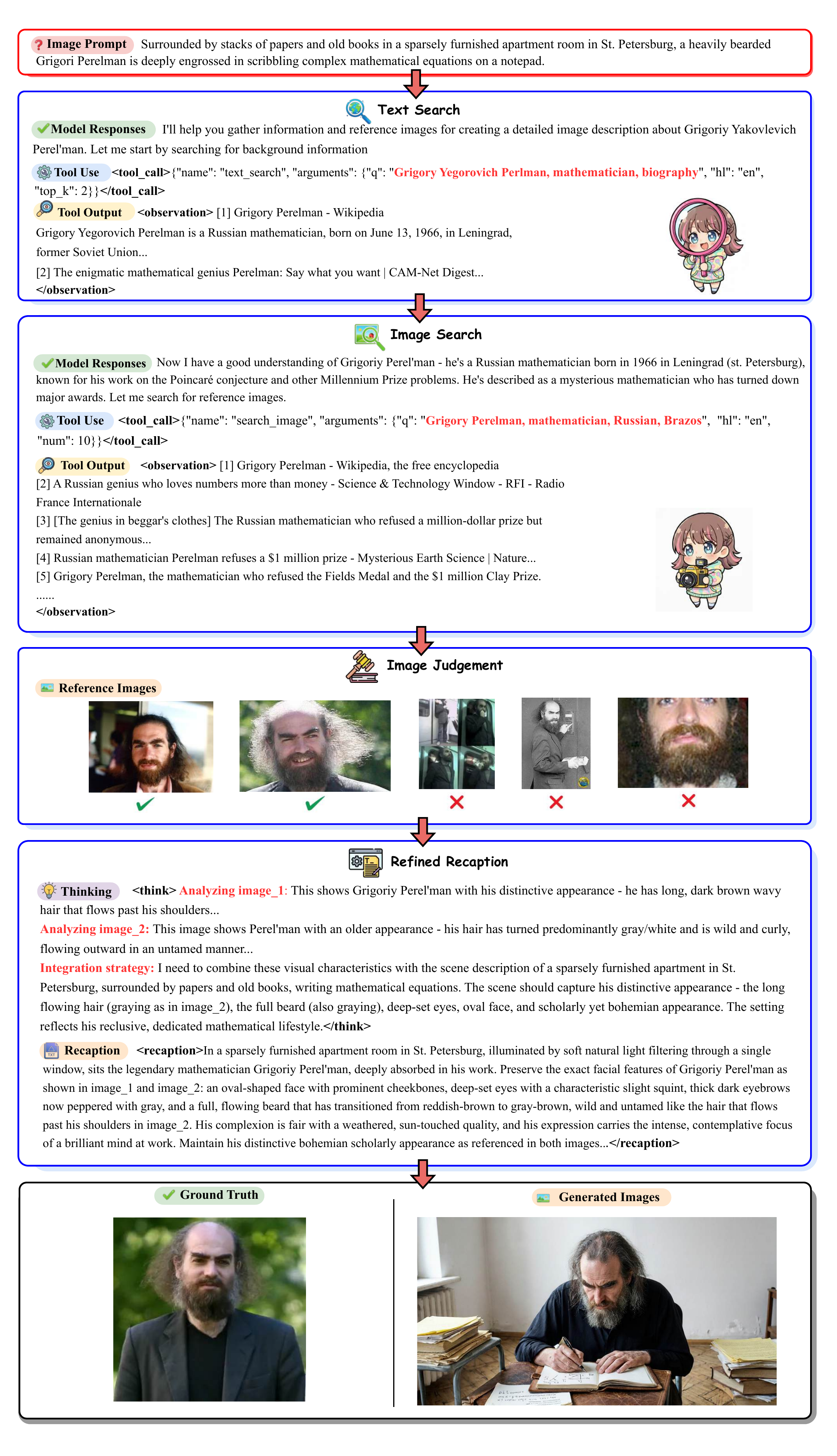}
    \vspace{-20pt}
    \captionsetup{justification=centering}
    \caption{Image generated for Grigory Perelman scribbling mathematical equations.}
    \label{fig:perelman}
    \vspace{-40pt}
\end{figure}

\paragraph{Case 3: Celebrity - William Butler Yeats}
This case details the generation of an image for a prompt depicting the poet \textbf{William Butler Yeats} in his rustic stone cottage study. The model first initiated a text search to gather precise biographical and physical descriptions, confirming iconic details such as his signature pince-nez glasses and historical background. It then performed a targeted image search to retrieve authentic historical portraits, capturing his distinct facial structure across different ages. By integrating these verified visual characteristics—such as his angular bone structure and tousled hair—into a refined recaption, the model was able to situate the historically accurate figure within the specific atmospheric setting of a manuscript-writing scene. See Fig. ~\ref{fig:yeets} for details.
\paragraph{Case 4: Celebrity - Grigory Perelman}
This case details the image generation process for the legendary mathematician Grigori Perelman. The model initially performed a text search to establish his reclusive persona and historical context in St. Petersburg, providing the thematic foundation for the "sparsely furnished apartment" scene. It then conducted a targeted image search to capture his highly distinctive "bohemian scholarly" appearance, characterized by wild, untamed hair and a thick, flowing beard. Through the image judgement and refined recaptioning stages, the model meticulously analyzed his facial features across different ages—specifically noting the transition from dark to gray-streaked hair—and integrated these traits with atmospheric details like scattered papers and soft natural light. This strategy ensured the final generation achieved profound biographical accuracy while vividly portraying the intense focus of a brilliant mind. See Fig. ~\ref{fig:perelman} for details.

\begin{figure}[!t]
    \centering
    \includegraphics[width=2.0\textwidth,height=1.0\textheight,keepaspectratio]{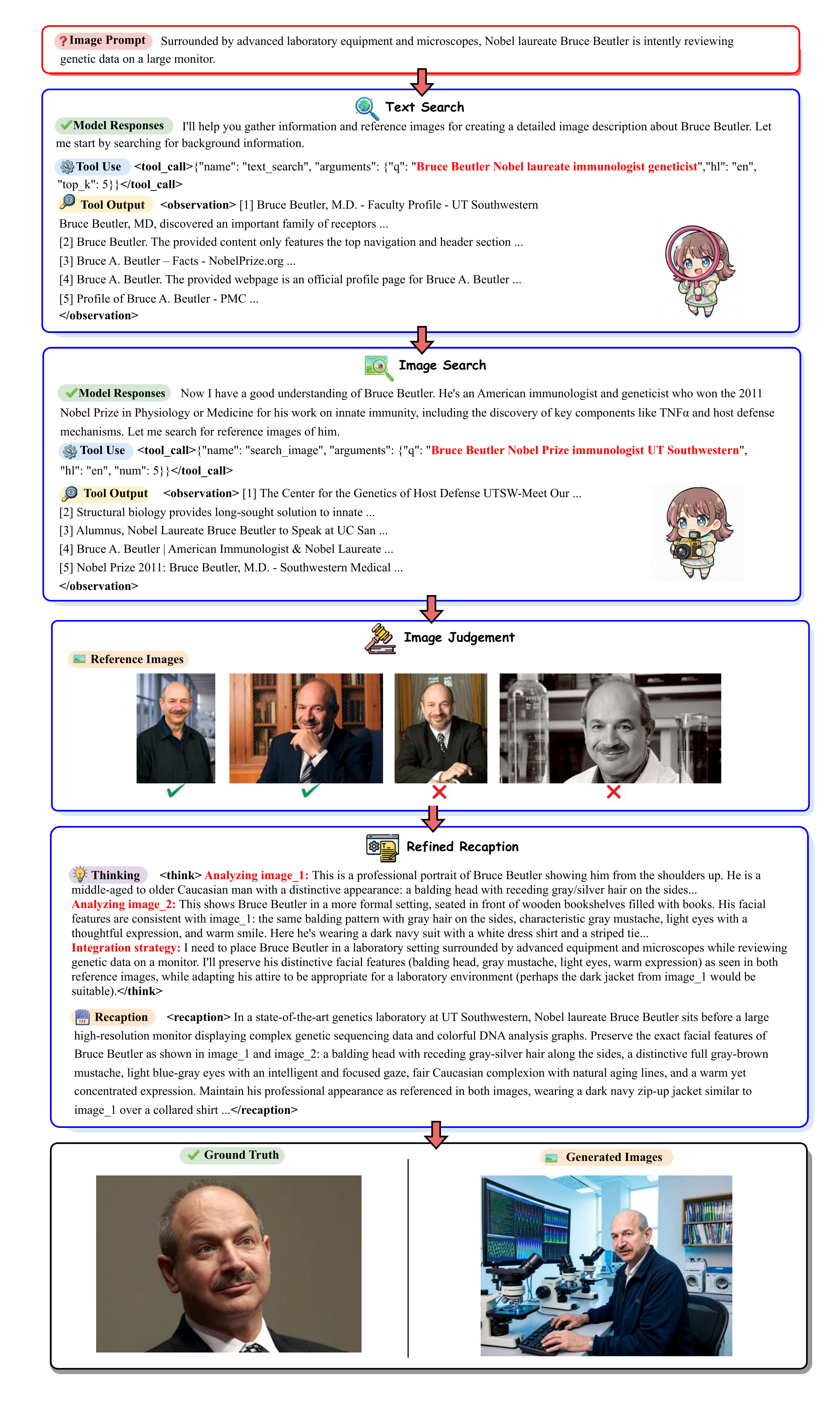}
    \vspace{-20pt}
    \captionsetup{justification=centering}
    \caption{Image generated for Bruce Beutler reviewing data on a monitor.}
    \label{fig:bruce}
    \vspace{-40pt}
\end{figure}

\paragraph{Case 5: Celebrity - Bruce Beutler}
This expert-focused case details the image generation process for \textbf{Nobel laureate Bruce Beutler in an advanced laboratory setting}. The model initially performed a text search to establish his professional identity as a Nobel Prize-winning immunologist and geneticist at UT Southwestern, ensuring the generated environment was contextually authentic. It then sourced reference portraits through image search and applied a rigorous image judgement process to filter for his most representative facial features, such as his receding gray-silver hair and distinctive mustache. By synthesizing these verified biographical details with the technical requirements of the scene in a refined recaption, the model achieved high accuracy in both persona and atmospheric realism. See Fig. ~\ref{fig:bruce} for details.

\newpage

\newpage

\newpage

\newpage

\newpage

\newpage
\section{FactIP Benchmark Evaluation Showcases}
\label{app: factip_showcases}
\begin{figure}[!t]
    \centering
    \includegraphics[width=1.0\textwidth,height=1.0\textheight,keepaspectratio]{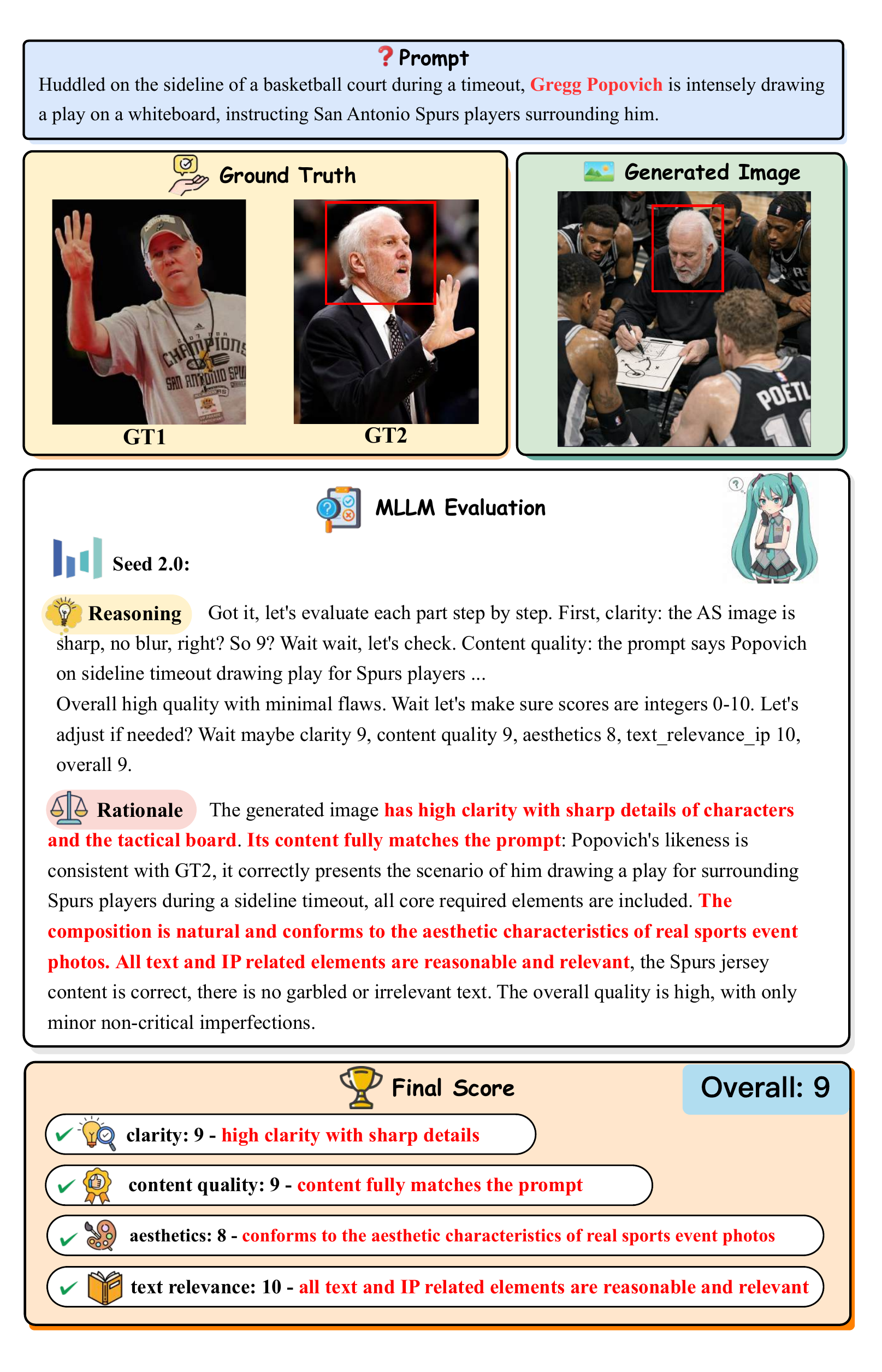}
    \vspace{-25pt}
    \captionsetup{justification=centering}
    \caption{MLLM evaluation for Popovich drawing a play.}
    \label{fig: Popovich}
    \vspace{-40pt}
\end{figure}

\paragraph{Case 1}
Figure~\ref{fig: Popovich} presents a representative Seed 2.0 evaluation example for an image generated from a prompt about Gregg Popovich. By comparing the synthesized result against the reference images (GT1 and GT2), the MLLM-based evaluator produces a fine-grained scoring profile together with an interpretable reasoning trace. In particular, the evaluator verifies both the facial resemblance to Popovich and the semantic consistency of basketball-related attributes, such as the Spurs jerseys and the tactical whiteboard. Based on this multi-aspect assessment, the sample receives a high overall score of 9, illustrating the benchmark’s ability to jointly measure identity fidelity and contextual relevance.

\begin{figure}[!t]
    \centering
    \includegraphics[width=1.0\textwidth,height=1.0\textheight,keepaspectratio]{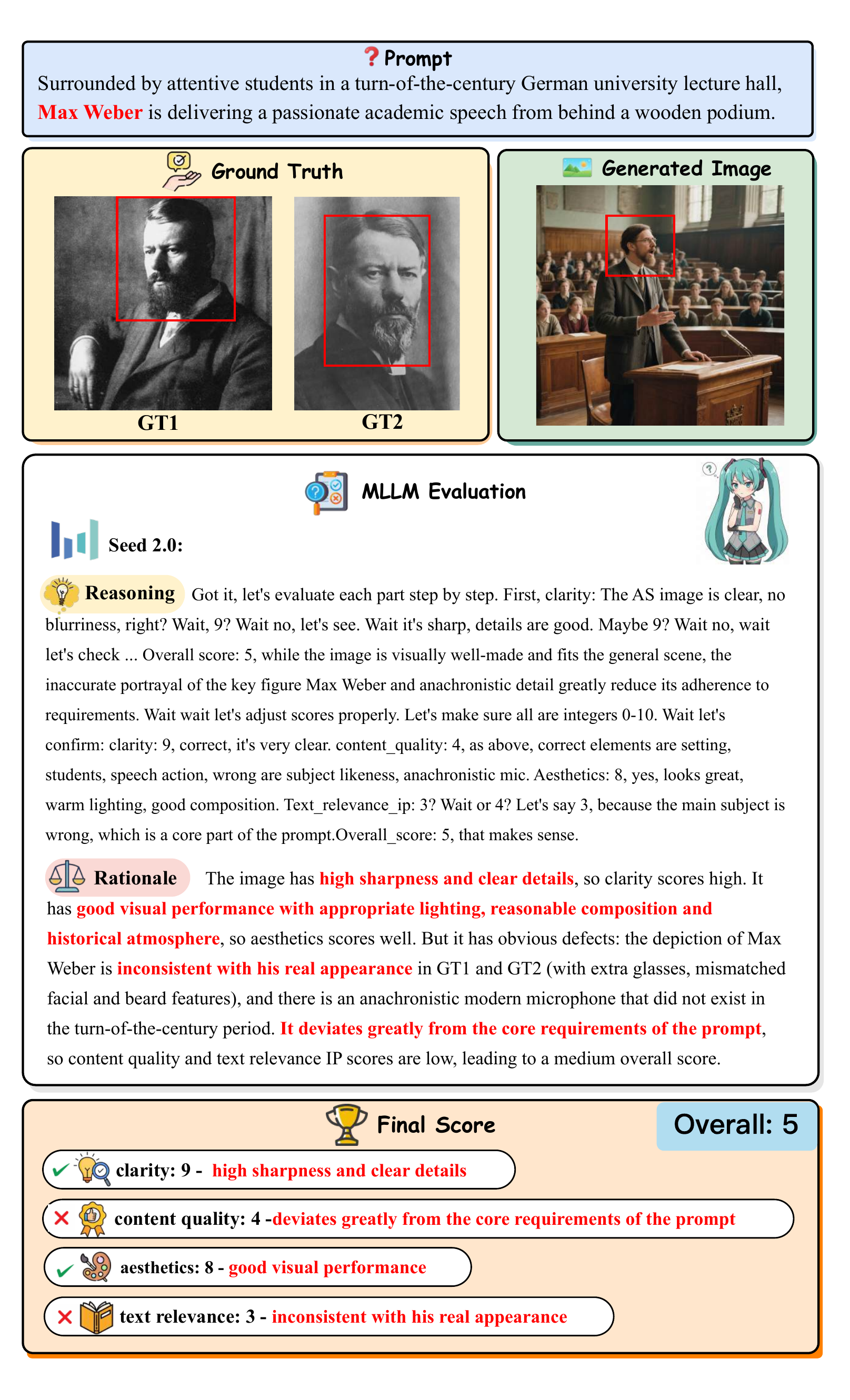}
    \vspace{-25pt}
    \captionsetup{justification=centering}
    \caption{MLLM evaluation for Max Weber delivering a passionate speech.}
    \label{fig: Weber}
    \vspace{-40pt}
\end{figure}

\paragraph{Case 2}
As shown in Fig.~\ref{fig: Weber}, this example illustrates how Seed 2.0 handles cases where visual quality remains strong but semantic correctness breaks down. For an image intended to depict Max Weber, the evaluator assigns only a moderate overall score of 5. Although the image is rated highly in terms of clarity (9) and aesthetic appeal (8), the reasoning process identifies two decisive issues: the generated face does not faithfully match the reference appearance---most notably due to the beard discrepancy and the addition of glasses absent from GT1/GT2---and the scene includes a modern microphone that introduces a clear historical anachronism. This case highlights that the benchmark does not merely reward surface-level image quality, but can also capture subtle factual and identity-level inconsistencies.

\begin{figure}[!t]
    \centering
    \includegraphics[width=1.0\textwidth,height=1.0\textheight,keepaspectratio]{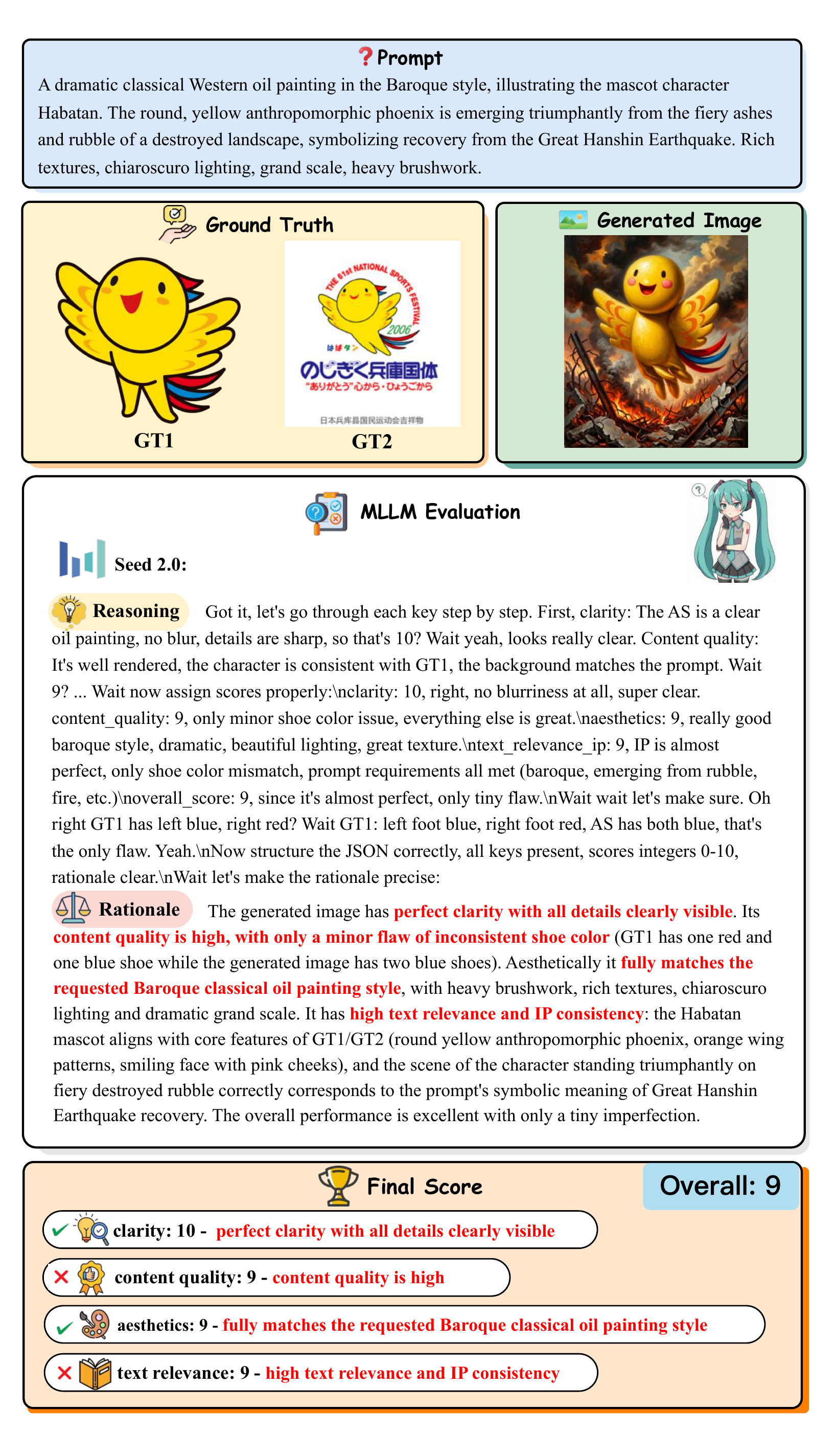}
    \vspace{-25pt}
    \captionsetup{justification=centering}
    \caption{MLLM evaluation for the mascot character Habatan in the Baroque style.}
    \label{fig: Habatan}
    \vspace{-40pt}
\end{figure}

\paragraph{Case 3}
Figure~\ref{fig: Habatan} provides an example of successful IP-oriented generation under a stylized artistic setting. Given a prompt requesting Habatan rendered in a Baroque oil painting style, the Seed 2.0 evaluator assigns an overall score of 9, reflecting both strong character fidelity and convincing stylistic execution. The accompanying reasoning demonstrates that the evaluator can recognize sophisticated visual properties associated with the Baroque tradition, including dramatic chiaroscuro and dense painterly brushwork, while still preserving the defining attributes of the mascot. Importantly, the analysis also uncovers a subtle but meaningful deviation: the generated image depicts two blue shoes, whereas the ground-truth reference (GT1) specifies one red shoe and one blue shoe. This example underscores the benchmark’s capacity for rigorous fine-grained auditing even in highly artistic generations.

\begin{figure}[!t]
    \centering
    \includegraphics[width=1.0\textwidth,height=1.0\textheight,keepaspectratio]{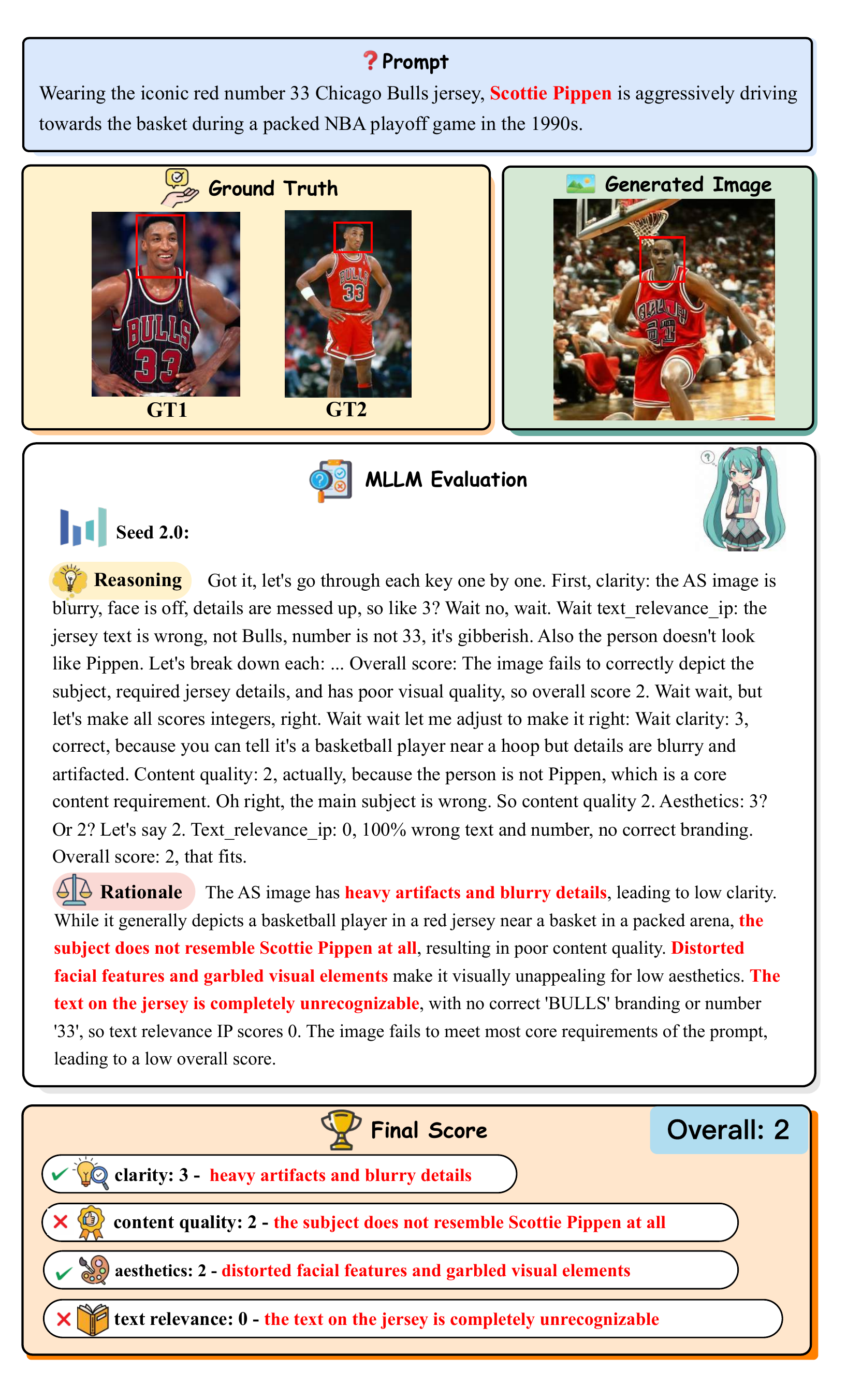}
    \vspace{-25pt}
    \captionsetup{justification=centering}
    \caption{MLLM evaluation for Scottie Pippen during an NBA playoff game.}
    \label{fig: Pippen}
    \vspace{-40pt}
\end{figure}

\paragraph{Case 4}
The failure mode illustrated in Fig.~\ref{fig: Pippen} further reveals the strictness of Seed 2.0 in evaluating identity-specific and text-sensitive attributes. For a prompt requiring Scottie Pippen wearing a classic Chicago Bulls jersey, the evaluator assigns a notably low overall score of 2. The reasoning trace points to several critical deficiencies: the generated subject bears little resemblance to Scottie Pippen when compared with GT1 and GT2, the image quality is compromised by substantial visual artifacts, and the jersey details---including the team branding and the iconic number 33---are rendered incorrectly and are largely illegible. This case demonstrates that the benchmark can effectively penalize failures in character identity, visual integrity, and fine-grained textual faithfulness.

\newpage

\newpage

\newpage

\newpage

\end{document}